%% file: main.tex
\documentclass[11pt]{article}

\usepackage[preprint]{acl}   
\usepackage{times}
\usepackage{latexsym}
\usepackage[T1]{fontenc}
\usepackage[utf8]{inputenc}
\usepackage{microtype}
\usepackage{inconsolata}
\usepackage{graphicx}
\usepackage{booktabs}
\usepackage{multirow}
\usepackage{amsmath}
\usepackage{amssymb}
\usepackage{subcaption}

\graphicspath{{figs_final/}}

\hypersetup{
  pdftitle={SIRF: A Spec-Internalized Risk Foundation Model for Industrial Content Risk Control},
  pdfauthor={Suwan Wu, Yumeng Lin, Pengcheng Yuan, Xiaolong Jiang},
  pdfsubject={EMNLP 2026 Industry Track},
  pdfkeywords={content risk control, spec internalization, continued pretraining, high-precision recall}
}

\title{SIRF: A Spec-Internalized Risk Foundation Model for Industrial Content Risk Control}

\author{
  Suwan Wu$^{1}$\thanks{These authors contributed equally.}, \quad Yumeng Lin$^{1,2}$\footnotemark[1], \quad Pengcheng Yuan$^{1}$, \quad Xiaolong Jiang$^{1}$ \\
  $^{1}$Xiaohongshu Inc. \qquad $^{2}$Tianjin University \\
  \texttt{\{wusuwan, linyumeng, yuanpengcheng, laige\}@xiaohongshu.com} \\
  \texttt{lym619@tju.edu.cn}
}

\begin{document}
\maketitle

\begingroup
\renewcommand{\thefootnote}{}
\footnotetext{Accepted at the Industry Track of the 2026 Conference on Empirical Methods in Natural Language Processing (EMNLP 2026).}
\endgroup
\addtocounter{footnote}{0}

\begin{abstract}
For industrial content risk control, the real deployment constraint is not average accuracy
but how much risk can be auto-handled under high precision and second-level latency. We present
\textbf{SIRF (Spec-Internalized Risk Foundation Model)}, which internalizes a platform's complex
policies, synthesized without additional human annotation via EntiGraph, MAGA rewriting and
account-level chain-of-thought (CoT), into the weights via continued pretraining (CPT), so rules
are applied at high precision under an ultra-low-latency, verdict-only deployment. A controlled same-source comparison
(Qwen3-8B-SFT vs.\ SIRF-8B-SFT, identical policy injection and verdict-only output form,
differing only in policy-grounded CPT) attributes the gain to internalization: SIRF-8B-SFT
reaches 71.3 Black Recall@P95, $+15.1$pp over the baseline, using only $\sim$70M CPT tokens
without harming general ability, and among included, logprob-available models under this
interface it matches or exceeds far larger systems. SIRF is deployed as a tree-model
adjudication layer ($20\%$ more mis-penalized samples recovered) and transfers to a freezing
scenario at low cost ($\sim$70\% relative mis-penalization reduction).
\end{abstract}

\section{Introduction}
\label{sec:intro}

\paragraph{Why is ``accuracy'' not enough for risk control?}
Content risk control is a high-stakes decision setting: a false positive wrongly penalizes an
innocent user and directly reduces that user's experience and activity, while a false negative
leaves a safety hazard, and both are irreversible. The objective is therefore not to ``make fewer
mistakes'' on average but to minimize disturbance to good users while protecting the ecosystem (a
motivation we validate with the online A/B result in \S\ref{sec:freezing}) --- that is, how much
risk can be auto-handled under \emph{high precision}. Yet account-level risk identification is
hard: the specifications are complex (around a hundred policies), the features are heterogeneous and
multi-source (profile, recent posts, comment interactions, devices, reports), and a verdict must be
returned within seconds from a full account view. Business strategies also change frequently, so
operators need to retune the threshold rather than retrain, which requires a monotone confidence
score. Under the joint constraints of high precision, second-level latency, verdict-only
output and a tunable threshold, applying complex rules at high precision becomes the deployment
bottleneck.

\paragraph{Why do existing routes not fit?}
\emph{Injecting the policy at inference} (long-context or retrieval-augmented generation, RAG)
incurs context and retrieval
overhead that conflicts with the second-level, single-forward constraint. \emph{Dropping the
policy and learning a pure classifier} would require well-covered samples for every policy and
every trigger--exemption branch, a labeling scale that is hard to meet for long-tail policies. In
both, rule knowledge either lives only transiently in the context or is never systematically
injected, leaving the rules outside the model's intrinsic capability.

\begin{figure}[t]
  \centering
  \includegraphics[width=\columnwidth]{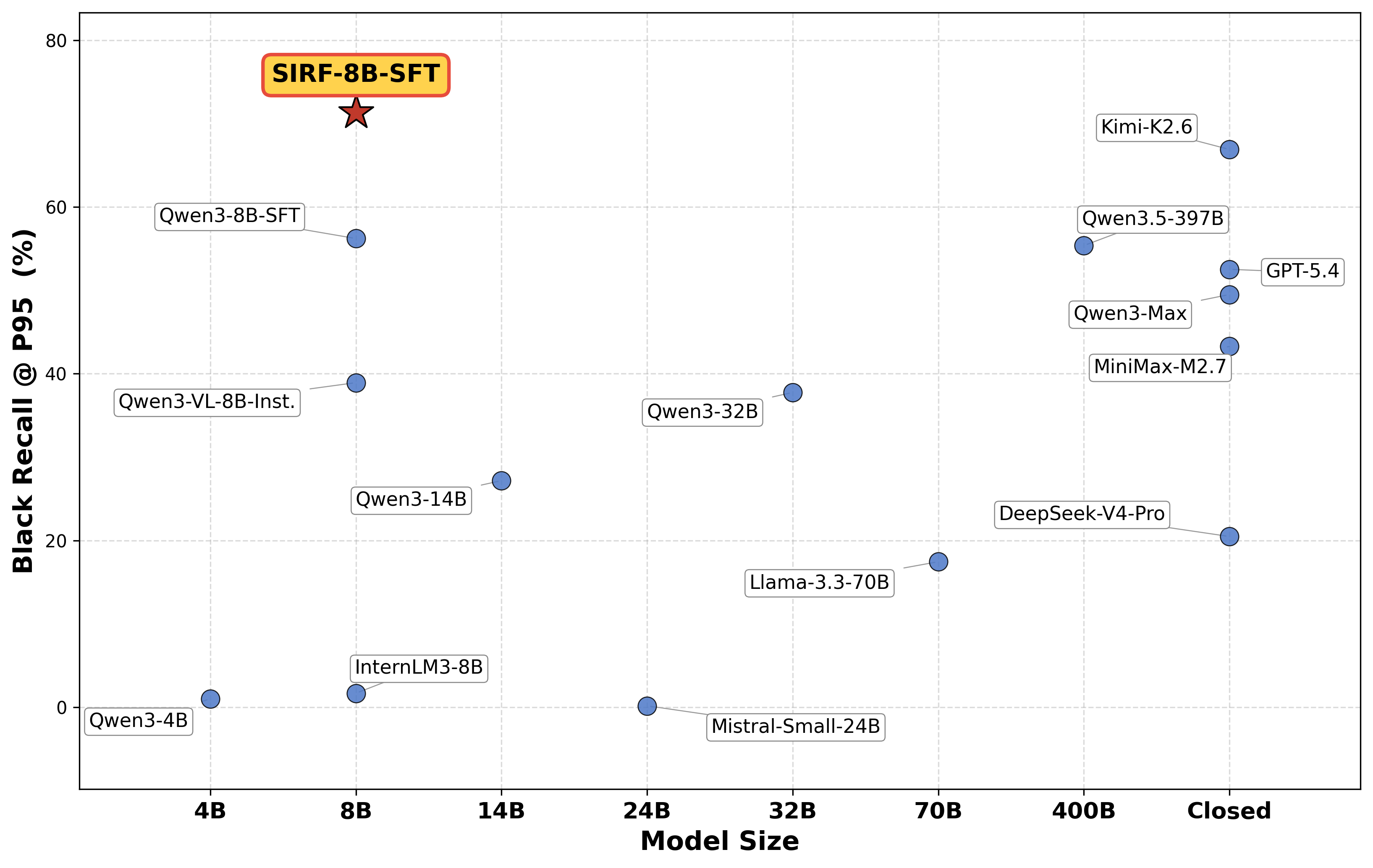}
  \caption{Black (risky) Recall@P95 vs.\ model size, among included, logprob-available models
  under this deployment interface. SIRF-8B sits above all compared open-weight (4B--400B) and
  closed-source models at the high-precision operating point. \textbf{Excluded models:} Claude,
  Doubao and GLM, whose APIs return no token logprobs, so no confidence ranking and hence no
  Recall@P can be computed; this is an interface limitation, not a capability judgment. Our core
  attribution to spec internalization does not rely on these cross-model comparisons but on the
  controlled 8B same-source study in Table~\ref{tab:main}.}
  \label{fig:teaser}
\end{figure}

\textbf{SIRF (Spec-Internalized Risk Foundation Model)} answers this by internalizing the policy
specifications (synthesized without additional human annotation via EntiGraph $+$ MAGA $+$ CoT)
into the weights via policy-grounded CPT, followed by domain supervised fine-tuning (SFT). The model then applies rules at
high precision under an ultra-low-latency, verdict-only deployment
(Figure~\ref{fig:teaser}). It is not merely a classifier but a transferable risk foundation model:
the same base, with light SFT, serves different risk domains.

\paragraph{Contributions.}
\begin{enumerate}\itemsep0pt
\item SIRF, a paradigm that internalizes risk-control policies into the weights with 70M
  synthesized tokens and no per-query retrieval.
\item A controlled attribution: a same-source comparison differing only in CPT, plus a
  per-component corpus ablation (\S\ref{sec:ablation}) isolating the gain to policy content rather
  than in-domain tokens.
\item Transferability: light SFT moves the base to a new risk domain ($\sim$70\% relative
  mis-penalization reduction).
\item Deployment evidence at second-level latency ($13\%$ prompt cut; $+20\%$ release), with general
  knowledge preserved and costs reported honestly.
\end{enumerate}

\section{Related Work}
\label{sec:related}

\paragraph{Synthetic CPT and spec internalization.}
Domain-adaptive pretraining (DAPT,~\citealp{gururangan2020dapt}) continues training on domain
corpora; EntiGraph~\citep{yang2024entigraph} synthesizes entity-relation text for synthetic CPT to
internalize domain facts. Closest to us is the concurrent Model Spec Midtraining
(MSM,~\citealp{li2026msm}), which midtrains on synthetic documents about a model spec to reduce
misalignment. SIRF shares this internalize-then-demonstrate paradigm but differs in goal: MSM
targets general alignment, whereas SIRF instantiates spec internalization for industrial
risk-control policies under strict deployment constraints, with a controlled attribution study,
online deployment and cross-domain transfer.

\paragraph{LLM data synthesis.}
Self-Instruct~\citep{wang2022selfinstruct} self-distills instruction data;
MAGA~\citep{hao2025maga} expands corpora via multi-genre, multi-audience rewriting. SIRF
specializes these for policy internalization, with label-leakage control.

\paragraph{Content moderation (guard models).}
Guard models from Llama Guard~\citep{inan2023llamaguard} and
ShieldGemma~\citep{zeng2024shieldgemma} to schema-conditioned
classification~\citep{zaratiana2026gliguard}, together with Constitutional
AI~\citep{bai2022constitutional} and RAG~\citep{lewis2020rag},
all supply the policy online (via prompt, schema or retrieval) and focus on classification quality.
SIRF instead internalizes it into the weights (no retrieval, verdict-only), targets the real
operating point, and gives deployment evidence for an adjudication layer and cross-domain
transfer.

\paragraph{Midtraining, data mixing, and schedule.}
Work on what to train on \emph{after} pretraining and \emph{before} alignment --- domain
reweighting~\citep{xie2023doremi}, late-training domain upsampling~\citep{blakeney2024spark} and
midtraining recipes in open models~\citep{olmo2024olmo2} --- finds late-stage gains sensitive to
mixture and schedule rather than token volume alone. SIRF instantiates this regime for a deployed
risk-control spec at a small budget; \S\ref{sec:ablation} is the corresponding
mixture-sensitivity check.

\paragraph{Selective classification.}
Abstaining below a confidence threshold is the classical error--reject
trade-off~\citep{chow1970reject}, formalized as selective classification with a coverage/risk
curve~\citep{elyaniv2010selective} and extended to deep
networks~\citep{geifman2017selective}. Recall@P reads this trade-off in the direction operators
care about: fix precision on the risky class, then ask how much of it can be auto-handled.

\paragraph{LLM confidence, calibration, and CoT faithfulness.}
LLMs' verbalized confidence is often overconfident and poorly
calibrated~\citep{xiong2024calibration,tian2023justask}, and CoT is often unfaithful, post-hoc
rationalizing a decided conclusion~\citep{turpin2023cot}; querying the class token probability
better reflects what the model knows~\citep{kadavath2022know}, and guard-model calibration
matters for deployment~\citep{liu2025calibration}. Post-hoc calibration (temperature and Platt
scaling~\citep{guo2017calibration}, isotonic regression~\citep{zadrozny2002transforming};
see also~\citealp{desai2020calibration}) improves probability quality, but any strictly \emph{monotone}
recalibrator leaves Recall@P unchanged (\S\ref{sec:deploy}, Appendix~\ref{app:calib}): a
re-thresholdable operating point needs a good \emph{ranking}, not a low ECE. Hence SIRF's use of
the verbalizer first-token probability as the decision score.

\input{body}

\end{document}

%% file: body.tex
\section{Method}
\label{sec:method}

\begin{figure}[t]
  \centering
  \includegraphics[width=\columnwidth]{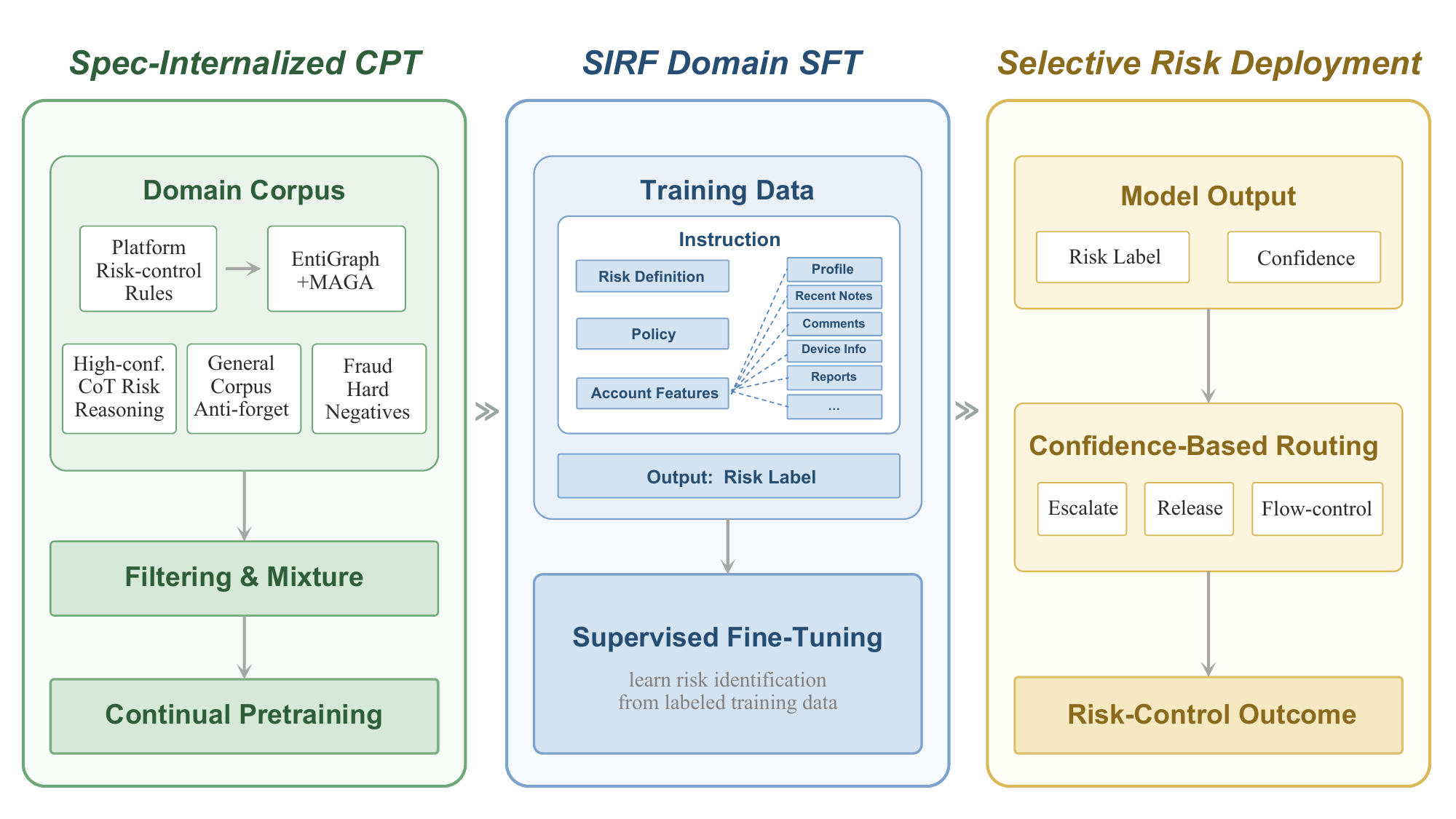}
  \caption{The SIRF pipeline: Spec-Internalized CPT $\rightarrow$ Domain SFT $\rightarrow$
  Selective Risk Deployment.}
  \label{fig:pipeline}
\end{figure}

SIRF's training pipeline is: \textbf{Base $\rightarrow$ policy-grounded CPT (internalizing
the policy specifications into the weights) $\rightarrow$ domain SFT $\rightarrow$
deployment (second-level, verdict-only)} (Figure~\ref{fig:pipeline}). We describe
corpus synthesis, training configuration, and deployment in turn.

\subsection{Policy Corpus Synthesis}
\label{sec:corpus}

We turn roughly a hundred structured policy specifications (each with trigger and exemption rules)
into a multi-perspective risk-control knowledge corpus by automated synthesis, with \textbf{no
additional human annotation}: an LLM pipeline builds the corpus from policy documents and account
features, commissioning no new labelling effort (scope of that claim below).

\paragraph{Decompose, and decouple stable rules from volatile strategies.}
Since rules change often and retraining each time is costly, we decompose the policies into
fine-grained underlying rules --- the slow-changing judgment logic, e.g.\ why a behavior is fraud ---
and separate them from the volatile strategy layer (whether a risk type is active, how tight its
threshold is). SIRF internalizes the stable rules into the CPT base and tracks the strategy layer via
dynamic thresholds (\S\ref{sec:deploy}) and light SFT, synthesizing along three lines.

\paragraph{(1) EntiGraph.}
For each policy we extract 6--15 key entities in 6 types (violating behavior, account feature,
threshold, concept, user type, exemption), describe each with its boundaries and confusable
distinctions, analyze cross-entity relations, generate trigger--exemption boundary samples, and
expand into 7 perspectives (case analysis, counterexamples, misjudgment review).

\paragraph{(2) MAGA rewriting.}
We rewrite each policy in 5 genres $\times$ 4 audiences (engineer, moderator, appealing user,
adversarial researcher) for expressive diversity.

\paragraph{(3) Account-level CoT distillation.}
We distill ``feature $\rightarrow$ policy-clause $\rightarrow$ decision'' chains from real
account-level data with a \emph{Self-Evolving Account Data Agent} (SEADA): a teacher LLM
(Qwen3.5-397B-A17B) turns account features plus the decomposed rules into a three-stage chain
(risk-signal check $\rightarrow$ exemption check $\rightarrow$ verdict); an independent
LLM-as-a-judge scores each chain for clause faithfulness; and a gate on trainability,
self-consistency and confidence admits only high-quality samples. Samples carrying
internal-ecosystem labels are removed, and no ground-truth disposition ever enters the data or the
teacher's context.

Outputs pass quality control (MD5 dedup, refusal detection, cleaning, label-balanced down-sampling);
generated text must be coherent, must not add rules beyond the policy, and must match the real
business form.

\paragraph{Scope of the annotation claim.}
The chains terminate in a verdict, so their conclusions act as a \emph{weak supervision signal}
during CPT: self-generated from features and rules, never conditioned on the ground-truth
disposition used for evaluation, and carried as fine-grained risk tags rather than the three-way
evaluation label. The claim is ``no \emph{new} human labels,'' not ``no supervision'' (see
Limitations).

\subsection{Training and Corpus}
\label{sec:train}

The CPT corpus totals $\sim$70M tokens ($\sim$$0.0002\%$ of Qwen3's $\sim$36T-token corpus;~\citealp{qwen3}): risk-reasoning
46.7M (66.7\%), structured policy knowledge 11.3M (16.1\%), anti-forgetting general corpus
6.8M (9.8\%) and public fraud hard negatives 5.2M (7.4\%); $\sim$58M directly internalize the
platform's rules. Token share overstates how much of the corpus is CoT: by \emph{document count}
the CoT block is only $22{,}674/101{,}669\approx22\%$, since one CoT document is far longer than a
policy-knowledge item. CPT uses next-token prediction for
1 epoch at a conservative LR ($1\text{e-}5$). The SFT
input is \texttt{\{policy\}$\backslash$n\{features\}} and the output is only the verdict, with
loss on response tokens.

\paragraph{Base checkpoint and baseline fairness.}
All three 8B models --- the zero-shot base, the SFT baseline (Qwen3-8B-SFT) and SIRF-8B-SFT ---
start from the \emph{identical} checkpoint \texttt{Qwen/Qwen3-VL-8B-Instruct}, used as a text-only
LLM with the vision tower frozen. The two fine-tuned arms share the same policy injection, the same
verdict-only output form and \textbf{exactly the same SFT data} ($40{,}989$ samples, same sources
and class balance, 3 epochs); the only difference is SIRF's added CPT stage, which is what makes
the gain attributable to it.

\subsection{Deployment}
\label{sec:deploy}

Online risk control needs second-level responses and only a verdict, so SIRF is deployed
\textbf{verdict-only}: the model emits the label and nothing else (4--5 tokens, no reasoning trace).
Rule application is internalized during CPT, so high-precision judgment needs no intermediate
reasoning, matching the SFT objective (no train/deploy gap).

\paragraph{Confidence: first-token verbalizer probability.}
Let $y_1$ be the first token the model emits. The decision score is $c=p(y_1\mid x)$, which under
greedy decoding equals $\max_v p(v\mid x)$ at that position. This one quantity is used throughout
the paper and online; Appendix~\ref{app:calib} compares it against an aggregate over label strings
that we do \emph{not} deploy. We avoid CoT confidence: the post-CoT class-token probability is shaped by the preceding
generation and CoT is often unfaithful~\citep{turpin2023cot}, while verbalized confidence is
generally overconfident~\citep{xiong2024calibration,tian2023justask}; the class token probability
better reflects the model's grasp~\citep{kadavath2022know}. The score lets operators retune without retraining; online we cut at an extreme percentile of the
score distribution (\S\ref{sec:flowctrl}). To avoid a clash of notation, ``P$x$'' always denotes a
\emph{precision} constraint in this paper, and percentile cutoffs are written as
$\tau_{q}$ ($q$ the percentile).

\paragraph{What the score must satisfy (and what it need not).}
Deployment needs not an absolutely calibrated probability but a \emph{ranking} from which a
high-precision operating point can be swept and re-thresholded without retraining. The distinction
matters: Recall@P is \emph{invariant} under any strictly monotone recalibration, so temperature or
Platt scaling cuts the expected calibration error (ECE) by $3$--$6\times$ in our setting while
leaving every operating point exactly where it was (Appendix~\ref{app:calib}). Hence: inference uses greedy decoding under a fixed
serving configuration, and thresholds are calibrated under it and fitted \emph{per model} on each
model's own confidence distribution, never transferred; and since a few fine-grained labels share
their first token, we record the first eight token probabilities and read the score at the first
position that disambiguates the predicted class.

\section{Experiments and Results}
\label{sec:exp}

\subsection{Experimental Setup}
\label{sec:setup}

\paragraph{Task and datasets.}
The task is account-level three-way risk classification (Black, risky; White, benign; Gray,
borderline) with 20$+$ fine subclasses. The main set ($n{=}1000$) follows the live production
distribution of a large content-community platform: \textbf{Black 596 / Gray 250 / White 154} over 25
fine classes (Appendix~\ref{app:evalset}), so the denominator behind Black Recall@P95 is $596$;
ground truth is the live disposition outcome plus a random human re-check. A balanced set ($n{=}4638$,
$\sim$200 per fine class) is a robustness check (\S\ref{sec:perclass}). Train/test split by time;
data are de-identified with internal-ecosystem labels filtered out.

\paragraph{Metrics.}
The core metric is per-class \textbf{Recall@P}, a one-vs-rest sweep run independently per class
rather than one global threshold: for class $k$ we take the \emph{smallest} score threshold
$t_k^{\star}$ at which the samples predicted $k$ with score $\ge t_k^{\star}$ reach precision $\ge P$,
and report the fraction of \emph{all} gold-$k$ samples it recovers (formal definition in
Appendix~\ref{app:evalset}).\footnote{E.g., Black Recall@P95 is the recall on risky accounts at a
threshold where Black precision is at least 95\%.} We focus on Black Recall@P90/P95, plus Macro-F1
and Accuracy; general ability uses 10 public benchmarks (\S\ref{sec:forget}) and efficiency uses
time-to-first-token (TTFT), end-to-end latency, throughput (QPS) and KV-cache memory
(\S\ref{sec:eff}).

\paragraph{Statistical reporting.}
A paired bootstrap with $1000$ resamples gives $\Delta$Black Recall@P95 $=+15.1$pp, 95\% CI
$[+11.7,+18.5]$, positive in $1000/1000$ resamples. Thresholds are fitted once on the full set and
held \emph{fixed} inside every resample, so the interval covers sampling variability of recall at a
fixed operating point but \emph{not} the variability of threshold selection. Two caveats follow.
Thresholds are selected on the same $n{=}1000$ set on which recall is reported, which is optimistic;
transferred unchanged to the independent balanced set they still hold Black precision $\ge95\%$
(SIRF $95.7\%$). And Recall@P95 is a step quantity at a steep cutoff, so a few boundary samples can
move it; we therefore rest not on a single P95 point but on the whole sweep
(Figure~\ref{fig:discrim}), the balanced-set replication (Appendix~\ref{app:balanced}) and the
multi-month deployment.

\paragraph{Compared models and fairness.}
We compare open-weight models (4B--400B) and closed-source models (GPT-5.4, Kimi-K2.6, Qwen3-Max,
MiniMax-M2.7, DeepSeek-V4-Pro). All see the same prompt template and policy injection, use greedy
decoding, and are thresholded on their own score distribution; the only interface difference is
GPT-5.4's \texttt{top\_logprobs} cap of 5 against 20 elsewhere. Models whose API returns no token
logprobs (\textbf{Claude, Doubao, GLM}) are excluded --- an interface limitation, not a capability
judgment. Granularity also differs: distinct first-token probability values are $17.5\%$ for GPT-5.4
versus $66$--$76\%$ elsewhere, so GPT-5.4's identical B@P90 and B@P95 reflect a coarse interface,
whereas for Kimi-K2.6 and Qwen3.5-397B the same pattern reflects a real ceiling on high-confidence
purity. Cross-model results corroborate but are not the central claim.

\subsection{Main Results}
\label{sec:main}

SIRF-8B-SFT reaches 71.3 Black Recall@P95, $+15.1$ points over the same-source, verdict-only
baseline Qwen3-8B-SFT (56.2), with Macro-F1 and Accuracy on par (Table~\ref{tab:main}).

\begin{table*}[!t]
  \centering
  \scriptsize
  \setlength{\tabcolsep}{4pt}
  \begin{tabular}{lrrrrrrrr}
    \toprule
    Model & B@P90 & \textbf{B@P95} & W@P90 & W@P95 & G@P90 & G@P95 & M-F1 & Acc \\
    \midrule
    GPT-5.4            & 52.5 & 52.5 & 6.5 & 3.9 & 0.4 & 0.4 & 63.7 & 65.7 \\
    Kimi-K2.6          & 66.9 & 66.9 & 27.2 & 25.2 & 0.0 & 0.0 & 70.0 & 73.3 \\
    Qwen3-Max          & 49.5 & 49.5 & 1.9 & 1.9 & 2.4 & 2.4 & 61.6 & 63.3 \\
    MiniMax-M2.7       & 55.2 & 43.3 & 0.6 & 0.6 & 10.8 & 6.0 & 63.8 & 65.4 \\
    DeepSeek-V4-Pro    & 31.5 & 20.5 & 6.5 & 6.5 & 7.6 & 2.0 & 42.1 & 41.5 \\
    Qwen3.5-397B-A17B  & 55.4 & 55.4 & 39.2 & \textbf{30.1} & 3.6 & 2.0 & 66.2 & 68.1 \\
    \midrule
    Qwen3-VL-8B-Inst.\ (0-shot) & 48.3 & 38.9 & 14.3 & 13.0 & 0.0 & 0.0 & 55.4 & 56.5 \\
    Qwen3-8B-SFT       & 83.4 & 56.2 & 46.8 & 27.3 & \textbf{32.8} & 18.0 & \textbf{81.9} & \textbf{84.0} \\
    \textbf{SIRF-8B-SFT}& \textbf{84.7} & \textbf{71.3} & \textbf{54.5} & 20.1 & 32.4 & \textbf{19.2} & 80.4 & 83.5 \\
    \bottomrule
  \end{tabular}
  \caption{Main results ($n{=}1000$, Recall@P \%). B/W/G $=$ Black/White/Gray, M-F1 $=$ Macro-F1;
  bold $=$ best in column. Upper block: closed-source APIs and very large open-weight models; lower
  block: the 8B same-source comparison, all three arms starting from the same
  \texttt{Qwen3-VL-8B-Instruct} checkpoint (\S\ref{sec:train}) and differing only in what training
  is applied. Only logprob-available models are included
  (\S\ref{sec:setup}). \emph{The central claim is the same-source
  Qwen3-8B-SFT vs.\ SIRF-8B-SFT comparison (differing only in CPT); cross-model numbers are
  corroboration under this interface, not a leaderboard.}}
  \label{tab:main}
\end{table*}

\begin{figure}[t]
  \centering
  \includegraphics[width=\columnwidth]{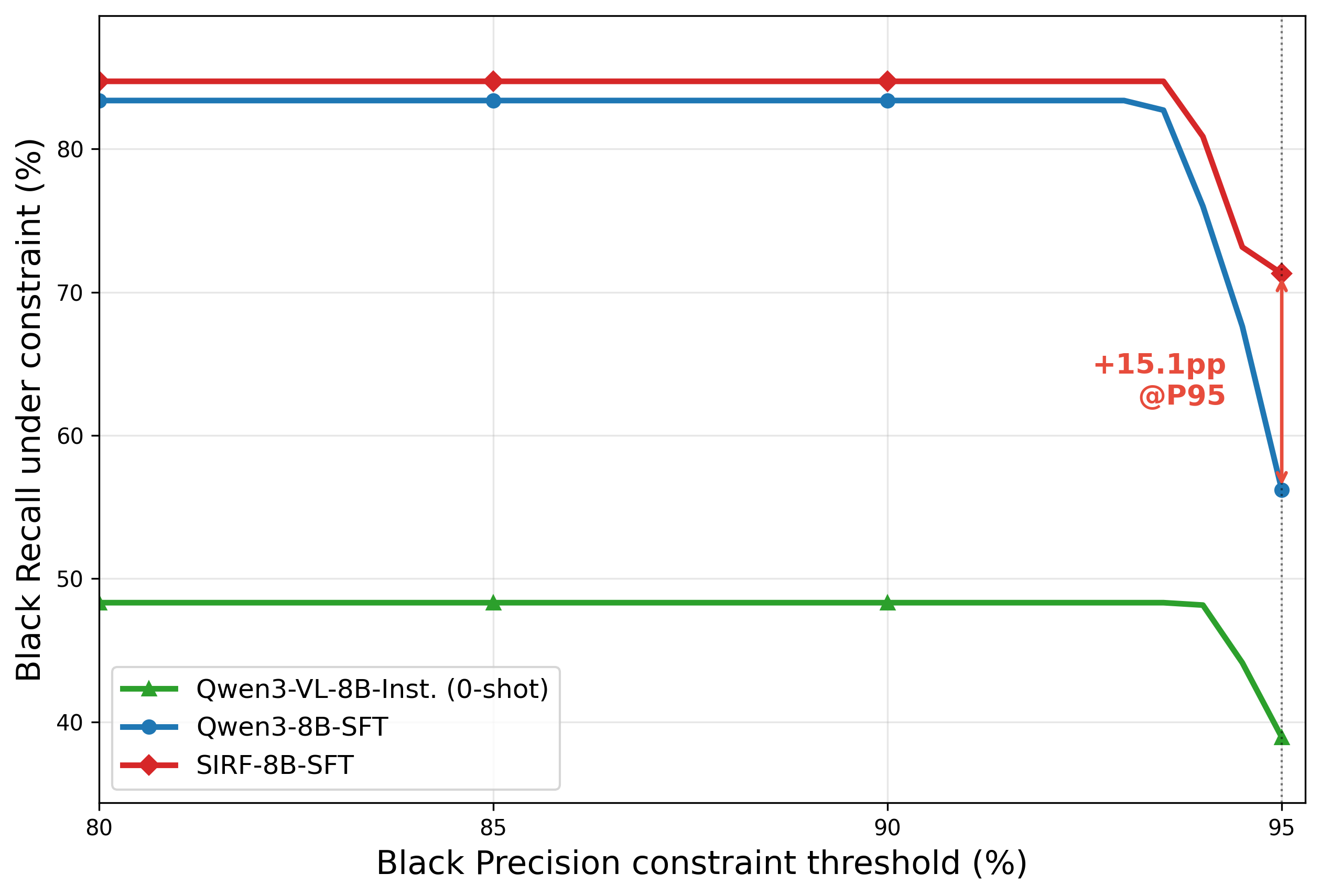}
  \caption{Black recall vs.\ precision constraint for the three same-source 8B arms
  (Qwen3-VL-8B-Inst.\ $\rightarrow$ Qwen3-8B-SFT $\rightarrow$ SIRF-8B-SFT); the gap is largest at
  P95 ($+15.1$pp).}
  \label{fig:discrim}
\end{figure}

\paragraph{Main findings.}
\emph{(i) Highest high-precision Black recall.} Black Recall@P95 reaches 71.3, $+15.1$pp over
Qwen3-8B-SFT and above every included logprob-available model --- Kimi-K2.6 at 66.9, the 400B
Qwen3.5-397B at 55.4 (Figure~\ref{fig:teaser}). \emph{(ii) The tighter the threshold, the larger the
advantage.} Black recall is flat over P80--P90 and only engages past $\sim$P93
(Figure~\ref{fig:discrim}): SIRF's high-confidence samples are purer and hold recall under stricter
precision, so the gap peaks at P95. \emph{(iii) The gain is from what was trained, not how much.}
70M is $\sim$$0.0002\%$ of pretraining at LR $1\text{e-}5$ and general ability is nearly unchanged
(Figure~\ref{fig:radar}); \S\ref{sec:ablation} shows policy-free in-domain tokens buy nothing, and
Logit Lens corroborates the mechanism (Appendix~\ref{app:lens}).

\emph{(iv) No loss of basic classification; the White regression is confined to the extreme cutoff.}
Macro-F1 (80.4 vs 81.9) and Accuracy (83.5 vs 84.0) are on par. White Recall@P95 drops (20.1 vs
27.3), in apparent tension with the release path of \S\ref{sec:flowctrl}; sweeping the White
operating point resolves this as a mismatch, not a conflict. At matched precision White recall is
$68.2$ vs $59.1$ at P85 and $54.5$ vs $46.8$ at P90 (White precision $85.4/85.0$, $90.3/90.0$), so
SIRF is the \emph{better} White judge across the release band and loses only at the steepest cutoff
--- the same right-shift mechanism as (ii). Human-review load does not grow, since SIRF acts only on the
two confident ends.

\paragraph{Confidence is higher and more usable as a threshold.}
SIRF's top-1 score distribution shifts right of the baseline's (mean $0.788{\to}0.814$), is higher
on $662/1000$ samples, and the lift concentrates where SIRF is correct rather than spreading as
indiscriminate overconfidence (Appendices~\ref{app:calib} and~\ref{app:conf}). This underpins
finding~(ii) and the extreme online cutoffs (\S\ref{sec:flowctrl}).

\subsection{Where Does the Gain Come From? Per-Component CPT Ablation}
\label{sec:ablation}

Table~\ref{tab:main} varies CPT as a single block, which supports ``CPT helps'' but not ``the spec
was internalized'': since account-level CoT dominates the corpus by token count, the gain could
equally come from distilled in-domain reasoning. We therefore replaced the CPT corpus component by
component, holding the base, the SFT data and schedule, and the evaluation set fixed. Black
Recall@P95 is $56.2$ with no CPT, $55.7$ for domain-only (fraud plus general corpus, no policy),
$64.6$ for EntiGraph$+$MAGA only, $69.5$ for CoT only, $70.6$ without CoT and $71.3$ for the full
corpus; across arms B@P90 stays within $81.2$--$84.7$ and Macro-F1 within $79.7$--$81.9$
(Appendix~\ref{app:ablation}).

This separates ``what was trained'' from ``how much''. \emph{In-domain tokens on their own buy
nothing}: domain-only is indistinguishable from no CPT, although that is exactly the arm the
``in-domain reasoning distillation'' reading predicts should capture most of the gain. They are not inert once policy
content is present (w/o CoT sits $6.0$pp above EntiGraph$+$MAGA alone), but that difference is within
the step-noise of the metric and we do not build on it. Both
policy-carrying arms improve over no CPT --- declarative (EntiGraph$+$MAGA) and procedural
(account-level CoT, whose chains ground each verdict in policy clauses) --- and the full corpus is
highest, though the ablation does not order the two carriers. A mechanism check agrees: the CPT-only
model (no SFT) emits policy codes such as ``P-0'' and cannot classify at all ($100\%$
non-classification first tokens), i.e., CPT installs policy knowledge and SFT the output routing
(Appendix~\ref{app:lens}).

\subsection{Inference Efficiency}
\label{sec:eff}

Internalization means the online prompt no longer needs the full policy to invoke the rules. We
deploy a conservative \textbf{reduced policy}: trimming redundant phrasing while keeping all rule
details, cutting the prompt by 13\% ($10.7$k$\to$$9.0$k characters of policy). B@P90
($84.7{\to}84.6$) and Accuracy ($83.5{\to}83.3$) are essentially unchanged, while the strictest
operating point costs $71.3{\to}66.9$ --- stated plainly rather than as slight, since $-4.4$pp is
about $30\%$ of the headline gain.

Since the KV footprint is almost entirely the prompt, this maps directly to high-concurrency
speedups: median end-to-end latency improves $\downarrow$8.8\%/$\downarrow$12.6\%/$\downarrow$18.2\%
at concurrency 16/32/100, QPS rises $+14\%$--$+23\%$ for concurrency $\geq$16, and per-request KV
cache drops $\sim$13\%, raising the maximum concurrency from 381 to 436
(Appendix~\ref{app:eff}).

\begin{figure}[t]
  \centering
  \begin{subfigure}[b]{0.33\columnwidth}
    \centering
    \includegraphics[width=\linewidth]{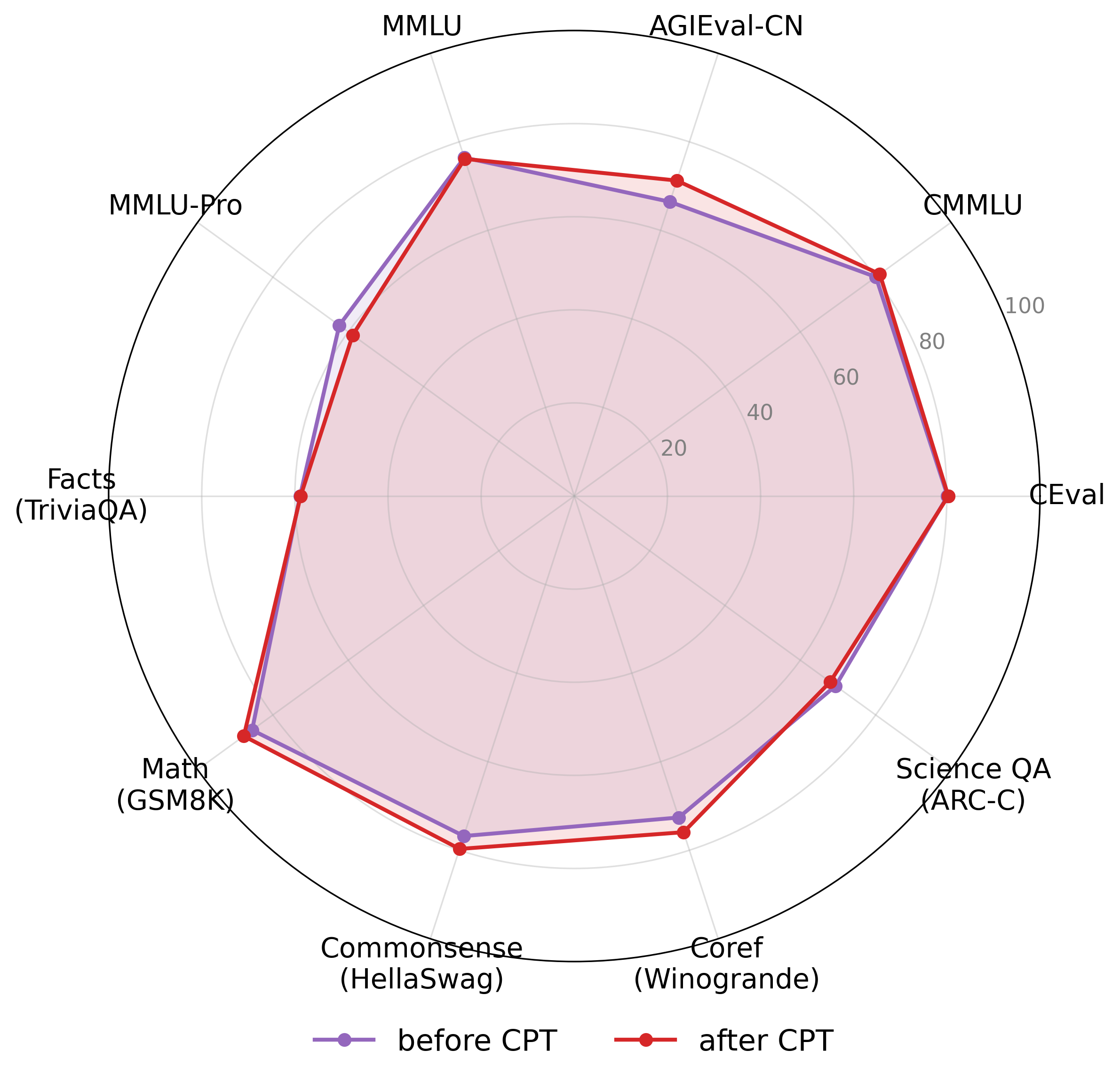}
    \caption{8B}
  \end{subfigure}\hfill
  \begin{subfigure}[b]{0.33\columnwidth}
    \centering
    \includegraphics[width=\linewidth]{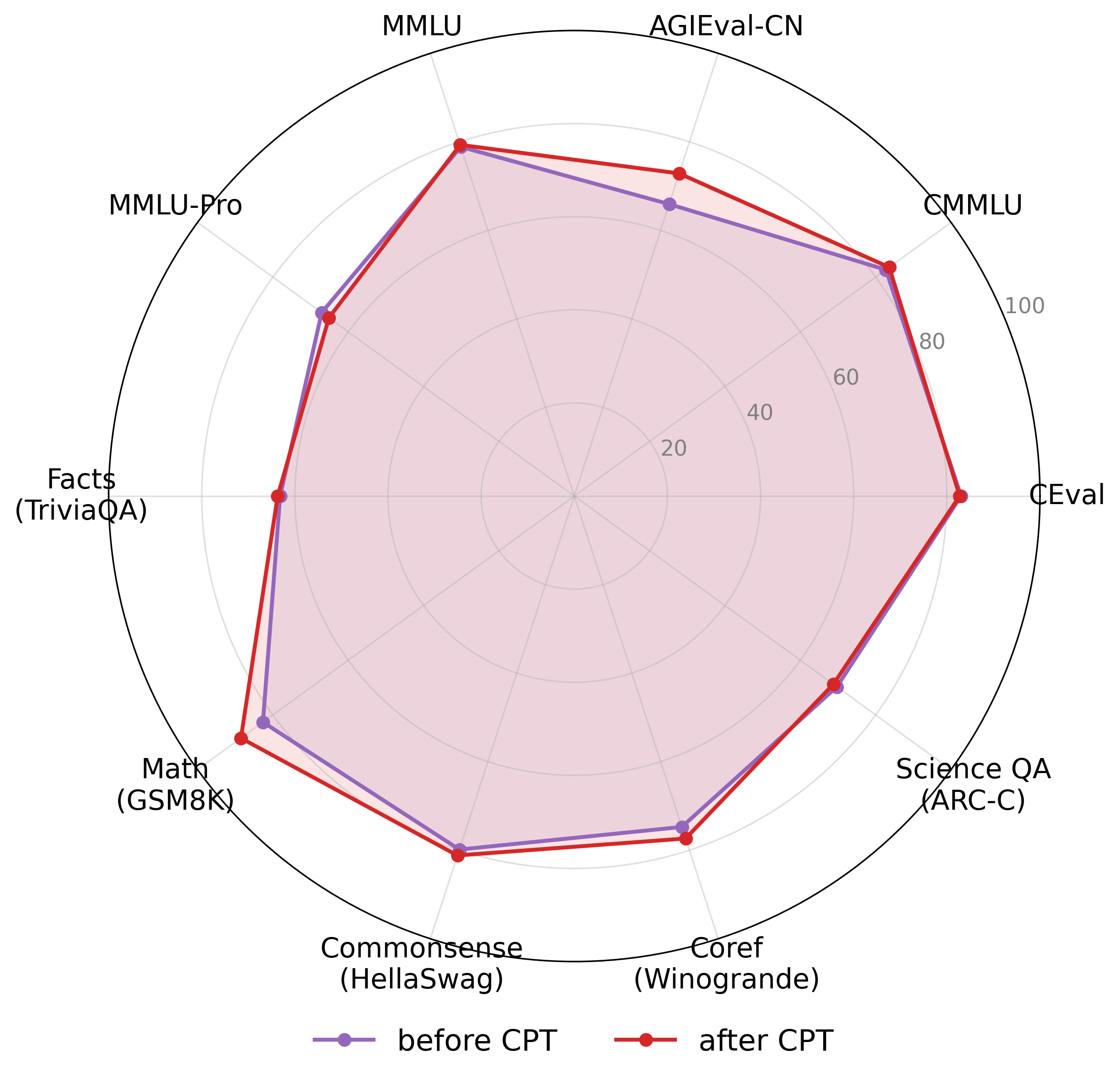}
    \caption{14B}
  \end{subfigure}\hfill
  \begin{subfigure}[b]{0.33\columnwidth}
    \centering
    \includegraphics[width=\linewidth]{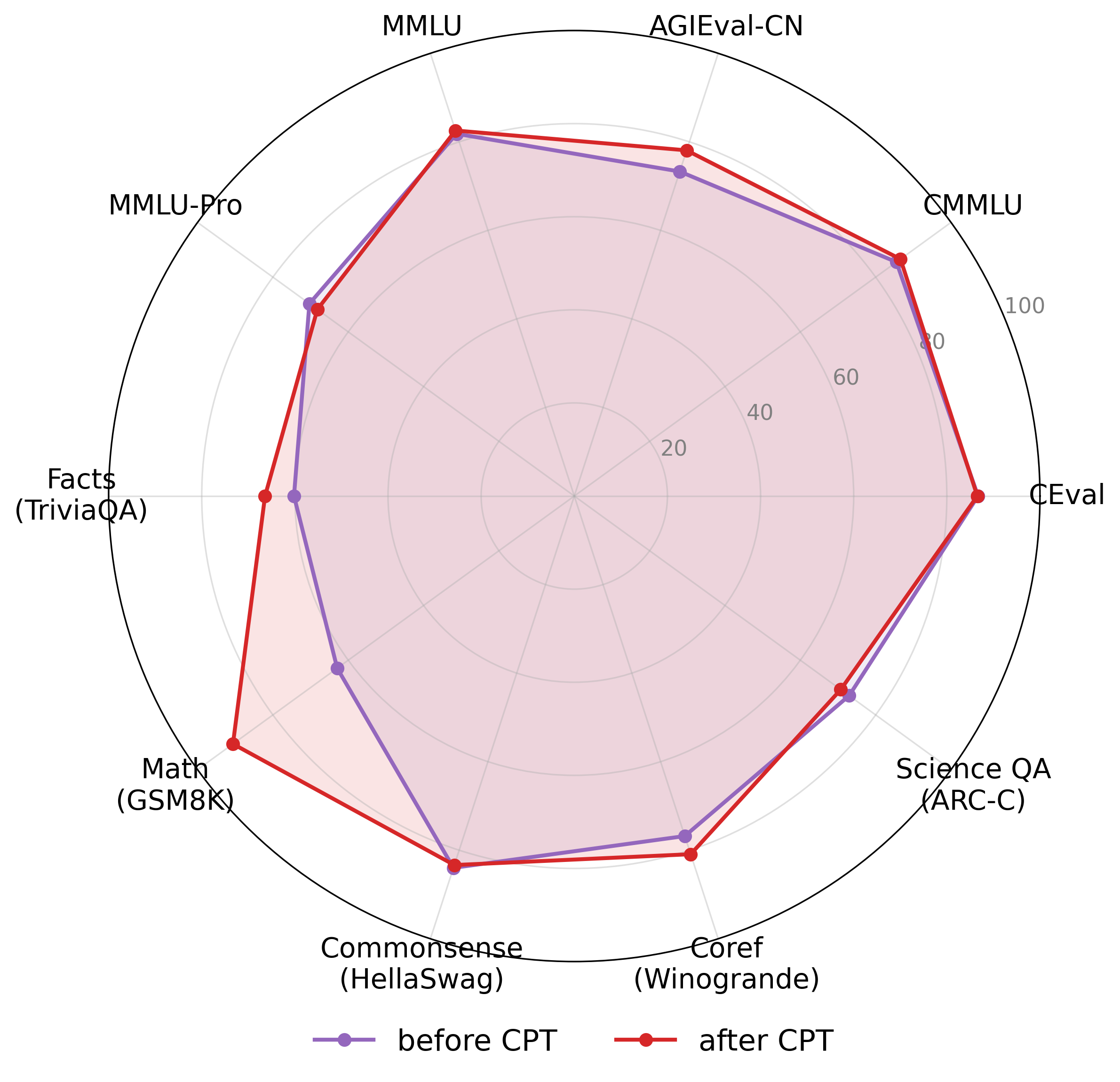}
    \caption{32B}
  \end{subfigure}
  \caption{Anti-forgetting radars at 8B, 14B, and 32B: general ability before/after CPT almost
  fully overlaps.}
  \label{fig:radar}
\end{figure}

\subsection{Per-Class Diagnosis}
\label{sec:perclass}

Deployment also depends on per-subclass behavior, which sets operators' disposition confidence per
risk class. On the balanced set's 24 fine classes, SIRF gains $\Delta\geq+3$ on 6, loses on 4 and
is flat on the rest, with class-average recall on par with the baseline; gains concentrate on
high-risk classes (infant conditioning $+8.8$, qualification documents $+6.6$, debt relief and fund
recovery $+6.5$) while drops are mostly low-risk types (per-class table in
Appendix~\ref{app:balanced}). With $\sim$200 samples per class the claim is consistency of
direction, not per-class significance: deltas of $+4$ to $+6.6$ are within sampling noise and only
the largest are individually meaningful.

\subsection{General-Knowledge Retention}
\label{sec:forget}

Does domain CPT harm general knowledge? We evaluate before/after CPT at 8B / 14B / 32B on 10 public
benchmarks: C-Eval~\citep{huang2023ceval}, CMMLU~\citep{li2024cmmlu},
AGIEval-CN~\citep{zhong2024agieval}, MMLU~\citep{hendrycks2021mmlu},
MMLU-Pro~\citep{wang2024mmlupro}, GSM8K~\citep{cobbe2021gsm8k},
HellaSwag~\citep{zellers2019hellaswag}, WinoGrande~\citep{sakaguchi2020winogrande},
ARC-Challenge~\citep{clark2018arc}, TriviaQA~\citep{joshi2017triviaqa}. The radars overlap almost
fully at every scale (Figure~\ref{fig:radar}): at 8B most benchmarks are flat or slightly up
(AGIEval-CN $+4.8$, GSM8K $+2.2$, HellaSwag $+2.9$) and only MMLU-Pro drops ($-3.6$); this holds at
14B / 32B. Targeted internalization at this budget
therefore costs little general ability, which is what makes the base transferable. Text-only CPT
does weaken instruction-following (see Limitations), which matters little for a verdict-only
deployment.

\section{Production Deployment Evidence}
\label{sec:business}

SIRF is deployed in two online domains: (a) flow control (SIRF as a tree-model adjudication layer)
and (b) freezing (cross-domain transfer), both running for multiple months with no degradation of
the high-precision gains. For compliance we report relative magnitudes, the order of magnitude of
affected traffic, the runtime in months and the randomization scheme, but \emph{not} absolute
volumes, business impact, exact thresholds, per-channel breakdowns or confidence intervals. This
section corroborates that the system works in production rather than measuring an effect size; the
quantitative claims rest on \S\ref{sec:exp}.

\subsection{Flow Control: Adjudication Layer}
\label{sec:flowctrl}

After a coarse hit by the tree model, SIRF adjudicates again (Figure~\ref{fig:pipeline}) under
ultra-conservative per-class cutoffs, routing hit traffic three ways --- auto-escalate
high-confidence Black, release high-confidence White, keep the rest under the original flow control.
Each cutoff is set on its own class's score distribution (Black at the $\tau_{99}$ percentile of the
\emph{Black} scores, White analogously), so the two auto-handled arms come from different
sub-populations, not from one top-$1\%$ slice of all hits. On hit traffic at the
hundred-thousand-user scale, auto-escalation covers about 10\% of all hits and release recovers good
samples equivalent to $+20\%$ of the tree model's original false positives.

\paragraph{Why Gray is never auto-disposed.}
Gray recall at strict precision is low for \emph{every} model in Table~\ref{tab:main} ($0$--$19.2$
at P95). This is a property of the class, not a defect: Gray is the ``undecided'' bucket, so its
score mass sits mid-distribution and no useful threshold exists. Gray therefore stays in the
original tree-model flow --- practitioners should route a Gray-like class, not threshold it.

\paragraph{Which White operating point the release path uses.}
The release arm needs high White precision and runs in the P85--P90 band, not at the P95 cutoff ---
exactly the band where SIRF beats the baseline on White recall at matched precision. That is the
mechanism behind the $+20\%$ release figure, and why the White Recall@P95 regression adds no
human-review volume: mid-confidence White cases were never in the auto-release path.

\subsection{Freezing: Cross-Domain Transfer}
\label{sec:freezing}

We transfer the same SIRF base to the freezing domain with only light in-domain SFT and no CPT
rerun: mis-penalization drops by about 70\% relatively, releasing a cumulative million-user scale
online.

An online A/B experiment validates these released users: a random split into treatment (SIRF
adjudication with high-confidence releases) and control (original policy) over the same population
and window, with no other model or policy change rolled out and no significant between-arm shift in
traffic mix, upstream thresholds or seasonal events. The treatment group is significantly better on
core metrics such as weekly active penetration (a relative lift in the high-single-digit to
low-double-digit range), showing SIRF recovers mis-frozen high-value active users.

This is a first validation of SIRF as a risk foundation model (multi-domain transfer is future
work), confirming the introduction's motivation: fewer false positives directly improve normal
users' experience and activity.

\section{Conclusion}
\label{sec:conclusion}

SIRF internalizes a platform's risk-control policies into the weights with only 70M synthesized
tokens, so that under second-level latency and verdict-only output it recalls more risk at the
high-precision operating point ($+15.1$pp Black Recall@P95 over the same-source baseline) while
preserving general knowledge, with a per-component ablation attributing the gain to policy content
rather than extra in-domain tokens. A spec-internalized base fits industrial risk control better
than pursuing average accuracy.

\section*{Limitations}

\paragraph{Sensitivity of the operating-point metric.}
Black Recall@P95 is a steep-cutoff quantity and is therefore sensitive to a few boundary samples at
$n{=}1000$. We consequently do not rest on that single point but on combined evidence: consistent
dominance across P80--P95, the fixed-threshold paired difference ($+15.1$pp, 95\% CI
$[+11.7,+18.5]$), the balanced-set replication (Appendix~\ref{app:balanced}) and the multi-month
deployment. The thresholds are also fitted and evaluated on the same set; transferred unchanged to
the independent balanced set they still hold Black precision $\ge95\%$. A larger, multi-period
evaluation set remains future work.

\paragraph{Supervision.}
CPT uses no additional human annotation, but the SEADA chains terminate in verdicts that act as a
self-generated, quality-gated weak supervision signal (\S\ref{sec:corpus}); the ground-truth
disposition never enters the data or the teacher's context.

\paragraph{The score is a ranking, not a probability.}
Recall@P is invariant to any strictly monotone recalibration, so absolute calibration was never the
design target; diagnostics are reported in Appendix~\ref{app:calib} for completeness. A post-hoc
calibrator, or a marginal aggregation over label strings, is an orthogonal enhancement rather than
a prerequisite for the deployed operating point.

\paragraph{Costs of the deployed stance.}
Text-only CPT preserves knowledge and reasoning but weakens instruction-following
(\S\ref{sec:forget}); this is tolerable because deployment emits a single verdict. The reduced
policy trades $-4.4$pp at P95 ($71.3{\to}66.9$, roughly $30\%$ of the same-source gain) for a
$13\%$ prompt cut.

\paragraph{Scope and reproducibility.}
Comparison coverage is interface-limited: Claude, Doubao and GLM return no token logprobs and could
not be included. All experiments use the policies of a single Chinese-language platform, and the
foundation-model claim rests on a first transfer scenario (\S\ref{sec:freezing}). The data are
proprietary and no code or model is released; we document the prompt structure and serialization
conventions instead (Appendix~\ref{app:prompt}).

\section*{Ethical Considerations}

Content risk control directly affects users' speech and account rights, and a misjudgment may
wrongly penalize a legitimate user, so the boundary and accountability of automated decisions
are especially important. SIRF adopts ultra-conservative thresholds online, concentrating
automatic execution on the samples the model is most confident about, to reduce
false-positive risk. On data, the account data used for training and evaluation are
anonymized and used only for risk assessment, with no individual profiling or secondary use.
We also note that policies and labels may carry the value judgments and preferences of a
specific platform, so the model may perform unevenly across populations or content types; when
transferring SIRF to other platforms or scenarios, its fairness should be re-examined and the
policy re-calibrated. Finally, a risk-control system faces the dual risks of over-enforcement
that harms legitimate users and under-blocking that harms the ecosystem; the trade-off
should be decided jointly by the specific business's value orientation and human oversight,
rather than left entirely to the model.

\section*{Acknowledgments}

We thank the anonymous reviewers for their constructive comments. We also thank our colleagues on
the risk-control engineering and operations teams for supporting the online deployment and the A/B
evaluation.

\bibliography{references}

\clearpage
\appendix

\section{Evaluation Protocol and Set Composition}
\label{app:evalset}

\paragraph{Formal definition of Recall@P.} Let $\hat{y}_i$ be the predicted label, $y_i$ the gold
label and $c_i$ the decision score (\S\ref{sec:deploy}) of sample $i$. For class $k$ and target
precision $P$, let $A_k(t)=\{i:\hat{y}_i=k,\;c_i\ge t\}$ be the accepted set at threshold $t$, with
precision $\mathrm{prec}_k(t)=|\{i\in A_k(t):y_i=k\}|/|A_k(t)|$. Then
\begin{align*}
t_k^{\star} &= \min\{t:\mathrm{prec}_k(t)\ge P\},\\[-1pt]
\text{Recall@}P(k) &= \frac{|\{i\in A_k(t_k^{\star}):y_i=k\}|}{|\{i:y_i=k\}|}.
\end{align*}
Taking the smallest feasible $t$ makes the accepted set as large as the precision constraint allows.
Each class is thresholded on its own score distribution, and each model on its own; no threshold is
shared across classes or transferred across models.

The main set ($n{=}1000$) is a production-distribution sample from a single time window. Its
three-way composition is \textbf{Black 596 / Gray 250 / White 154}, so the denominators of the
Recall@P columns in Table~\ref{tab:main} are $596$, $250$ and $154$ respectively; its ground truth is
the live disposition outcome plus a random human re-check. Under the mapping used throughout, the
single undisclosed-merchant subclass forms Gray, the benign class forms White, and the remaining 23
fine-grained subclasses form Black.

Because that production distribution is heavily long-tailed, the smaller Black subclasses carry too
few samples for per-subclass conclusions on this set, so we make none: all per-subclass analysis
(\S\ref{sec:perclass}) uses the balanced set instead, which holds $\sim$200 samples per fine class
by construction. The three-way supports above are what the headline metrics depend on, and they are
large enough for the Black and Gray columns; the White column rests on $154$ samples and should be
read with that in mind.

\section{Calibration and Confidence Variants}
\label{app:calib}

This appendix reports calibration diagnostics for the decision score and compares it against
equally cheap alternatives (Table~\ref{tab:calib}). The result is worth stating bluntly: SIRF is \emph{less} well calibrated
in absolute terms than its same-source baseline, and this does not affect the operating point.

\begin{table}[h]
  \centering
  \small
  \setlength{\tabcolsep}{4pt}
  \begin{tabular}{llrr}
    \toprule
    Score & Model & ECE & Brier \\
    \midrule
    \multirow{2}{*}{first-token prob.}
      & Qwen3-8B-SFT & 0.128 & 0.197 \\
      & SIRF-8B-SFT  & 0.176 & 0.214 \\
    \midrule
    \multirow{2}{*}{$+$ Platt scaling}
      & Qwen3-8B-SFT & 0.038 & --- \\
      & SIRF-8B-SFT  & 0.029 & --- \\
    \midrule
    \multirow{2}{*}{$+$ isotonic}
      & Qwen3-8B-SFT & 0.033 & --- \\
      & SIRF-8B-SFT  & 0.034 & --- \\
    \midrule
    \multirow{2}{*}{marginal over labels}
      & Qwen3-8B-SFT & 0.081 & 0.123 \\
      & SIRF-8B-SFT  & 0.086 & 0.131 \\
    \bottomrule
  \end{tabular}
  \caption{Calibration of the decision score ($n{=}1000$, 10 equal-width bins, correctness of
  the emitted fine-grained label). Post-hoc calibrators are fitted on one random half and
  evaluated on the other. Temperature scaling behaves like Platt scaling and is omitted.}
  \label{tab:calib}
\end{table}

Three findings. (i) \emph{SIRF is more over-confident}: its ECE rises from $0.128$ to $0.176$,
because CPT raises confidence across the board while accuracy stays flat. (ii) \emph{Monotone
recalibration is free but useless here}: Platt and temperature scaling reduce ECE by $3.4\times$ for
the baseline and $6.1\times$ for SIRF, yet they are strictly increasing functions of the raw score and
therefore leave Recall@P at every precision level \emph{exactly} unchanged --- a re-thresholdable operating
point depends on the ranking, not on the probability values, which is why we keep the raw score
online and why a low ECE was never the design target. (iii) \emph{The aggregation matters more
than the calibrator}: summing the first-token probability mass over all label strings of a class
(``marginal over labels'') is equally cheap, better calibrated, and changes the ranking, unlike
the monotone calibrators. We did not use it in the deployed system, whose thresholds were
calibrated on the first-token score, but we recommend it as the default for anyone reusing this
recipe.

\section{Balanced-Set Replication}
\label{app:balanced}

The balanced set ($n{=}4638$, $\sim$200 per fine class) is a complementary robustness check.
Table~\ref{tab:balanced} gives its numbers. The direction of the same-source comparison is reproduced (SIRF above Qwen3-8B-SFT, both
far above the zero-shot base) with a much smaller margin than on the production-distribution set,
and Recall@P90 equals Recall@P95 for every model here, i.e., the precision constraint is not
binding on this distribution. Thresholds fitted on the main set and transferred unchanged to this
set still hold Black precision $\geq95\%$ (SIRF $95.7\%$), which is the cross-dataset check
referenced in \S\ref{sec:setup}.

\begin{table}[h]
  \centering
  \small
  \begin{tabular}{lrr}
    \toprule
    Model & B@P90 & B@P95 \\
    \midrule
    Qwen3-VL-8B-Inst.\ (0-shot) & 42.0 & 42.0 \\
    Qwen3-8B-SFT  & 78.1 & 78.1 \\
    \textbf{SIRF-8B-SFT} & \textbf{79.5} & \textbf{79.5} \\
    SIRF-14B-SFT  & 78.8 & 78.8 \\
    SIRF-32B-SFT  & 79.1 & 79.1 \\
    \bottomrule
  \end{tabular}
  \caption{Black Recall@P on the balanced set ($n{=}4638$).}
  \label{tab:balanced}
\end{table}

Table~\ref{tab:perclass} lists the six most-improved fine classes referenced in
\S\ref{sec:perclass}.

\begin{table}[h]
  \centering
  \small
  \setlength{\tabcolsep}{3pt}
  \begin{tabular}{lrrrrrr}
    \toprule
    Model & C1 & C2 & C3 & C4 & C5 & C6 \\
    \midrule
    Qwen3-8B-SFT & 4.1 & 6.7 & 11.0 & 0.0 & 35.0 & 70.5 \\
    SIRF-8B-SFT  & 12.9 & 13.3 & 17.5 & 6.5 & 40.0 & 74.5 \\
    $\Delta$     & \textbf{+8.8} & \textbf{+6.6} & \textbf{+6.5} & \textbf{+6.5} & \textbf{+5.0} & \textbf{+4.0} \\
    \bottomrule
  \end{tabular}
  \caption{Per-class Recall@P95 on the balanced set, the 6 most-improved classes,
  $\Delta=$ SIRF $-$ baseline. C1--C6: infant conditioning, qualification documents, debt
  relief, fund recovery, counterfeit marketing, fortune-telling.}
  \label{tab:perclass}
\end{table}

\section{Confidence-Reliability Analysis}
\label{app:conf}

\begin{figure}[t]
  \centering
  \begin{subfigure}[b]{0.49\columnwidth}
    \centering
    \includegraphics[width=\linewidth,height=0.85\linewidth,keepaspectratio]{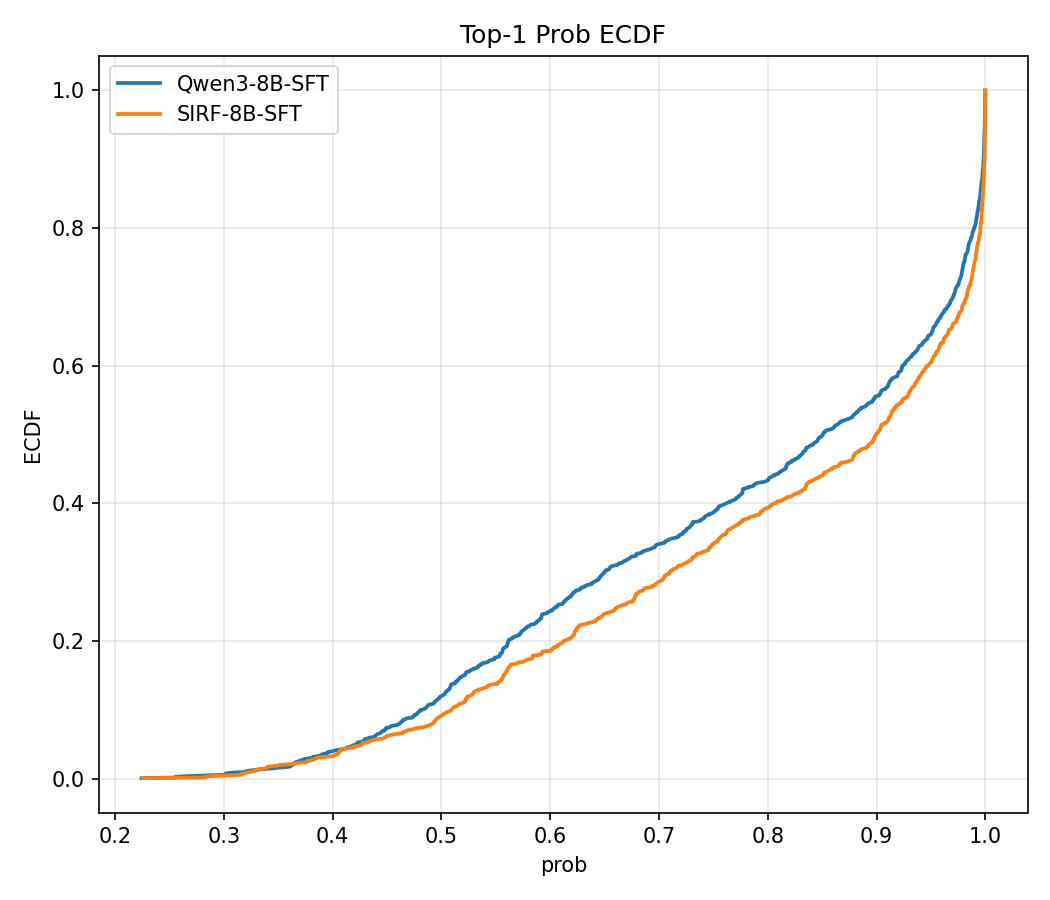}
    \caption{Top-1 probability ECDF.}
    \label{fig:confecdf}
  \end{subfigure}
  \hfill
  \begin{subfigure}[b]{0.49\columnwidth}
    \centering
    \includegraphics[width=\linewidth,height=0.85\linewidth,keepaspectratio]{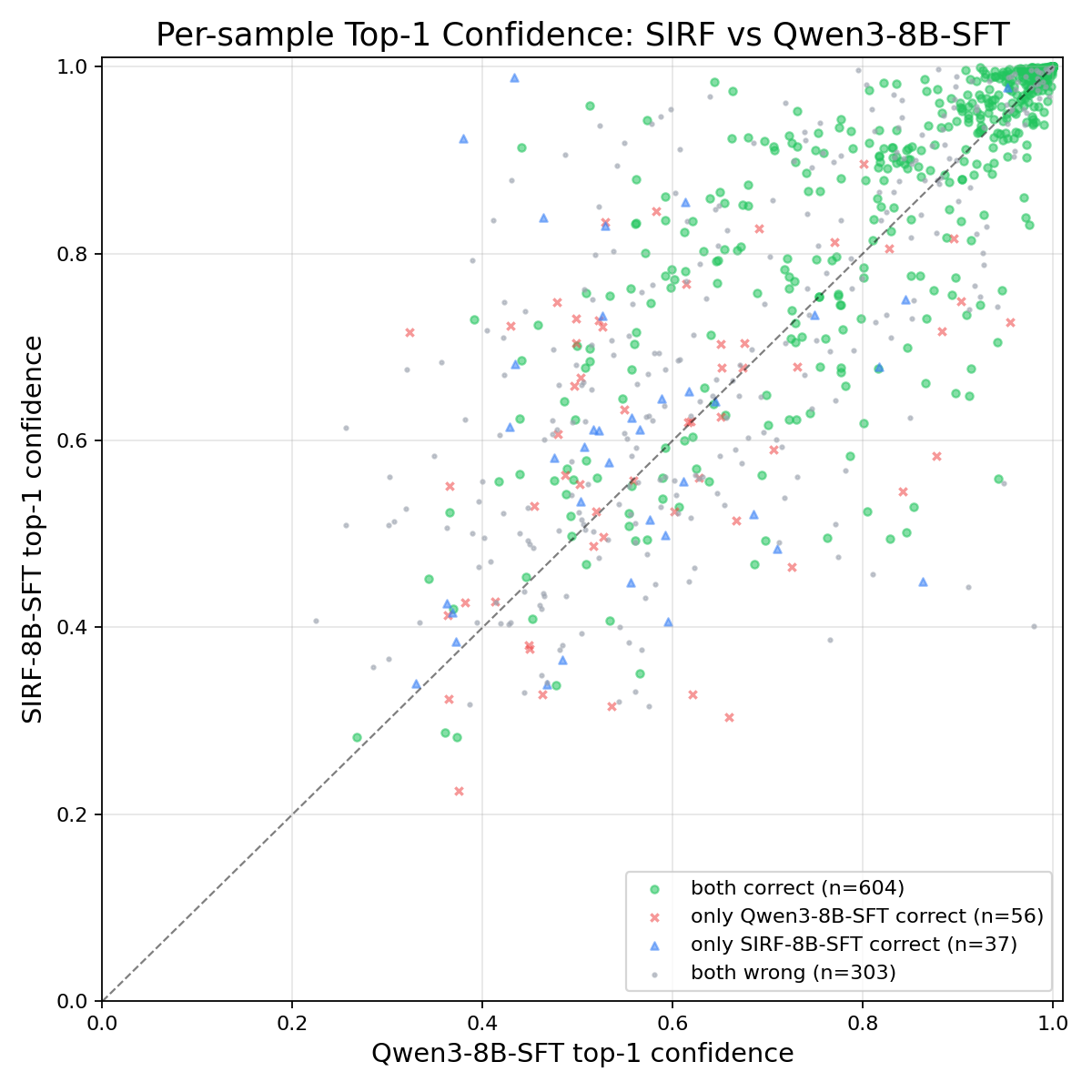}
    \caption{Per-sample scatter.}
    \label{fig:confscatter}
  \end{subfigure}
  \caption{First-token top-1 confidence: (a) SIRF's ECDF is right-shifted over the whole
  range; (b) per-sample scatter (quadrant-colored), most points above the diagonal.}
  \label{fig:conf}
\end{figure}

\begin{figure}[t]
  \centering
  \begin{subfigure}[b]{0.49\columnwidth}
    \centering
    \includegraphics[width=\linewidth,height=0.85\linewidth,keepaspectratio]{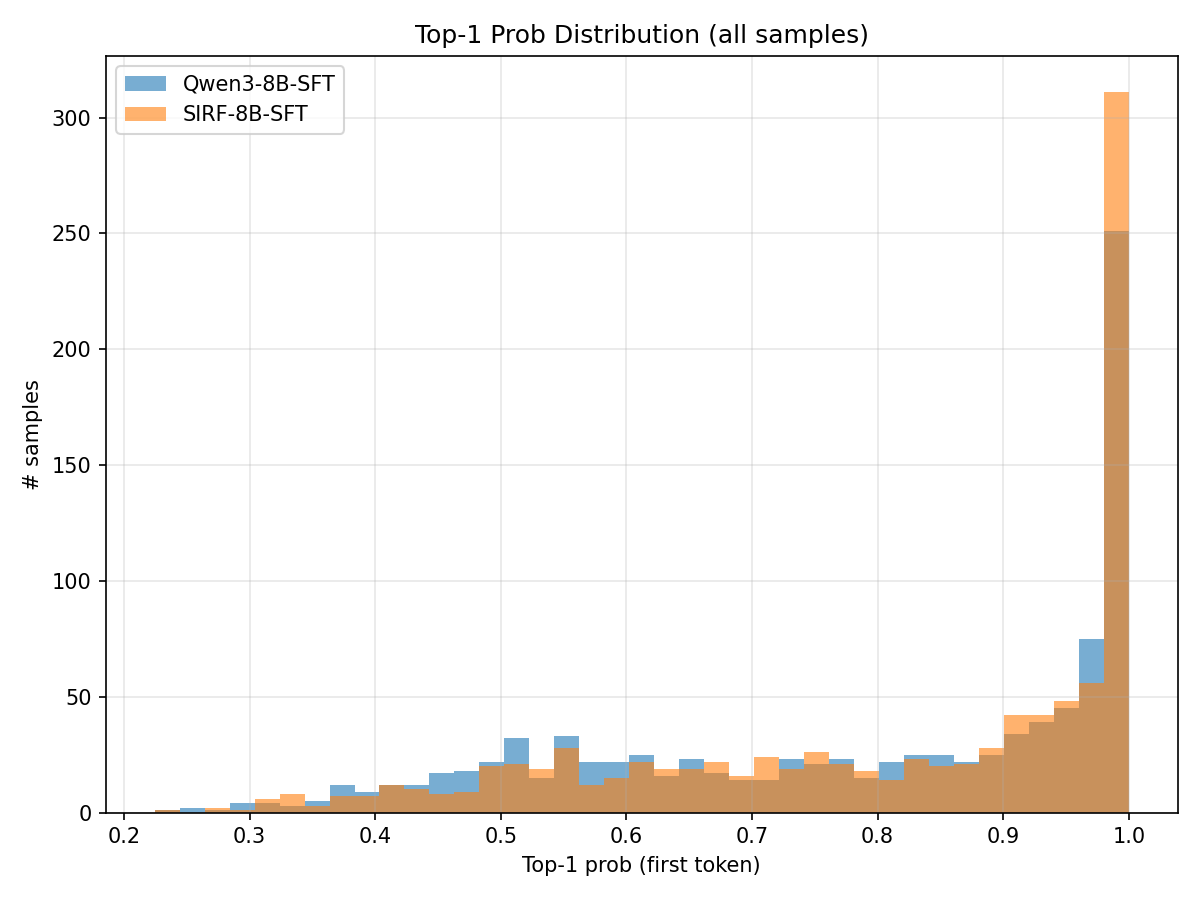}
    \caption{Top-1 probability histogram.}
    \label{fig:confdist}
  \end{subfigure}
  \hfill
  \begin{subfigure}[b]{0.49\columnwidth}
    \centering
    \includegraphics[width=\linewidth,height=0.85\linewidth,keepaspectratio]{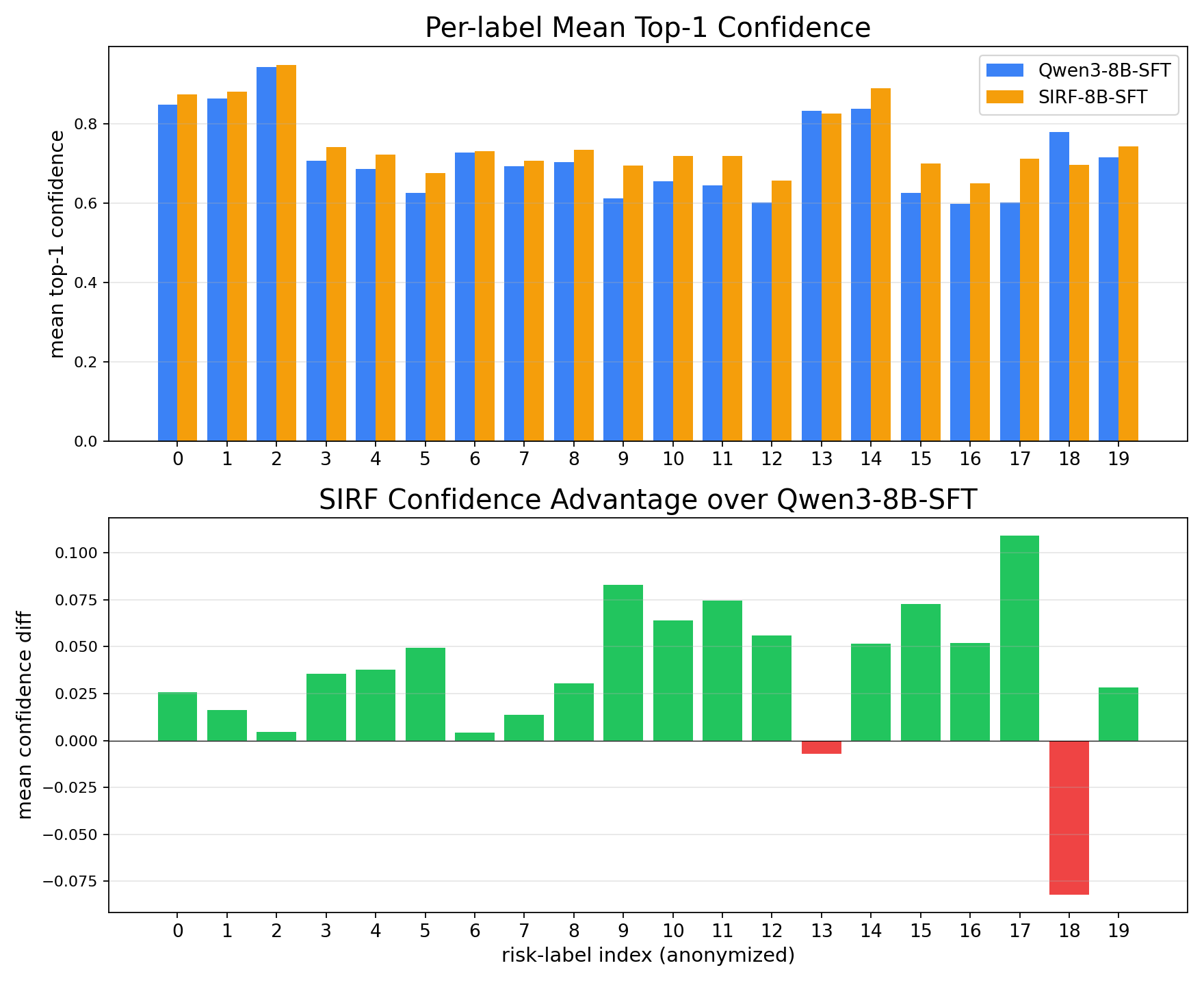}
    \caption{Per-label mean confidence.}
    \label{fig:perlabel}
  \end{subfigure}
  \caption{(a) SIRF has more mass in high-confidence bins; (b) per-label mean top-1
  confidence with SIRF's advantage consistent across labels.}
  \label{fig:confextra}
\end{figure}

This appendix gives the full first-token confidence analysis summarized in \S\ref{sec:main}.
Against the same-source baseline Qwen3-8B-SFT, we observe four consistent signals. (i)
\emph{Overall right-shift}: SIRF's top-1 probability mean $0.788{\to}0.814$, median
$0.851{\to}0.900$, with the empirical cumulative distribution function (ECDF) to the right over the
whole range
(Figure~\ref{fig:confecdf}) and more mass in the high-confidence bins
(Figure~\ref{fig:confdist}). (ii) \emph{Per-sample dominance}: SIRF exceeds the baseline on
$662/1000$ samples, with most points above the diagonal (Figure~\ref{fig:confscatter}).
(iii) \emph{The lift aligns with reliability}: it is largest where SIRF alone is correct
($+0.045$) and modest when both are correct ($+0.020$), not indiscriminate overconfidence.
(iv) \emph{Cross-label consistency}: SIRF's mean confidence exceeds the baseline on most
high-frequency labels (Figure~\ref{fig:perlabel}), with only a very few classes flat or
slightly lower. Together these underpin finding~2 and the extreme percentile cutoff used online (\S\ref{sec:flowctrl}).

\section{Efficiency Curves}
\label{app:eff}

\begin{figure}[t]
  \centering
  \includegraphics[width=\columnwidth]{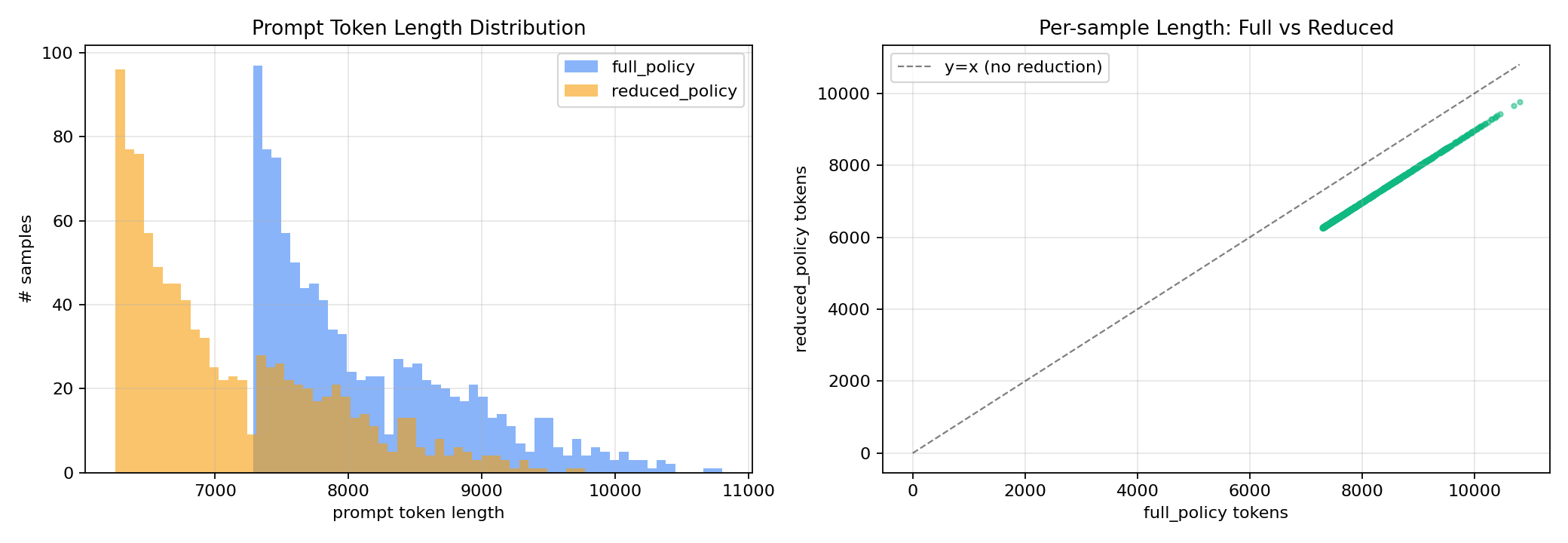}
  \caption{Prompt token-length distribution (full vs.\ reduced): the distributions have the
  same shape and only shift left, i.e., reduction is an approximately proportional trim per
  prompt.}
  \label{fig:a1}
\end{figure}

\begin{figure}[t]
  \centering
  \begin{subfigure}[b]{0.49\columnwidth}
    \centering
    \includegraphics[width=\linewidth,height=0.78\linewidth,keepaspectratio]{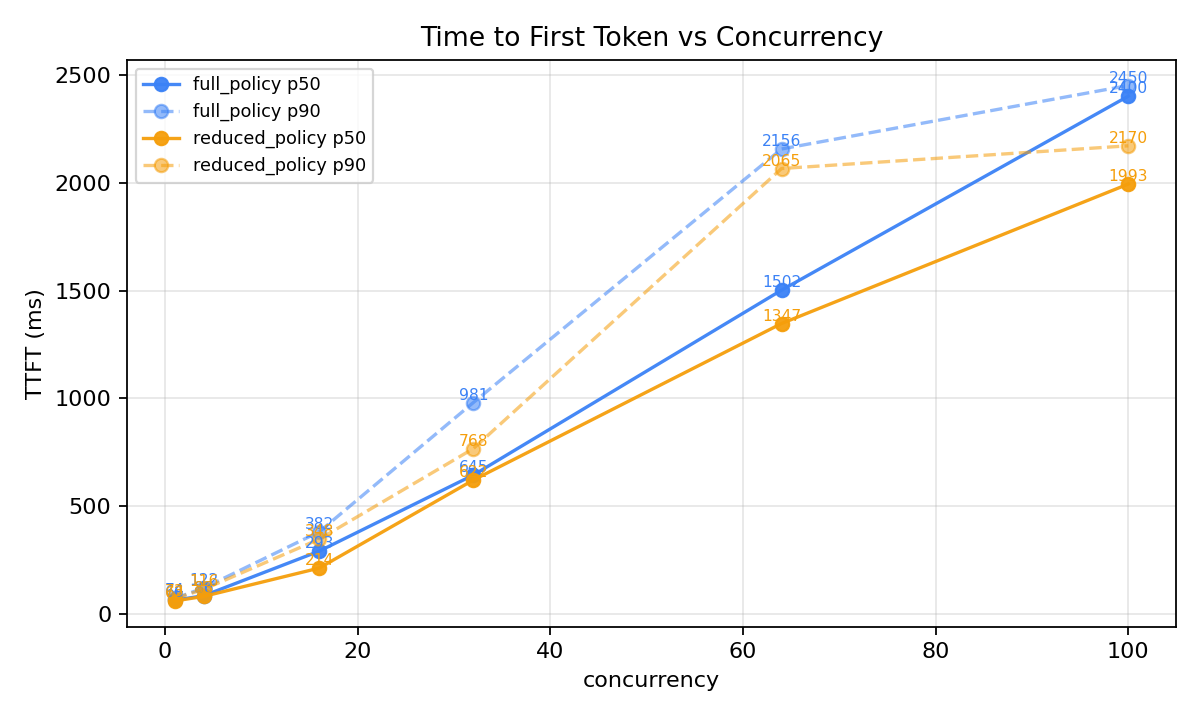}
    \caption{TTFT vs.\ concurrency.}
    \label{fig:a2}
  \end{subfigure}
  \hfill
  \begin{subfigure}[b]{0.49\columnwidth}
    \centering
    \includegraphics[width=\linewidth,height=0.78\linewidth,keepaspectratio]{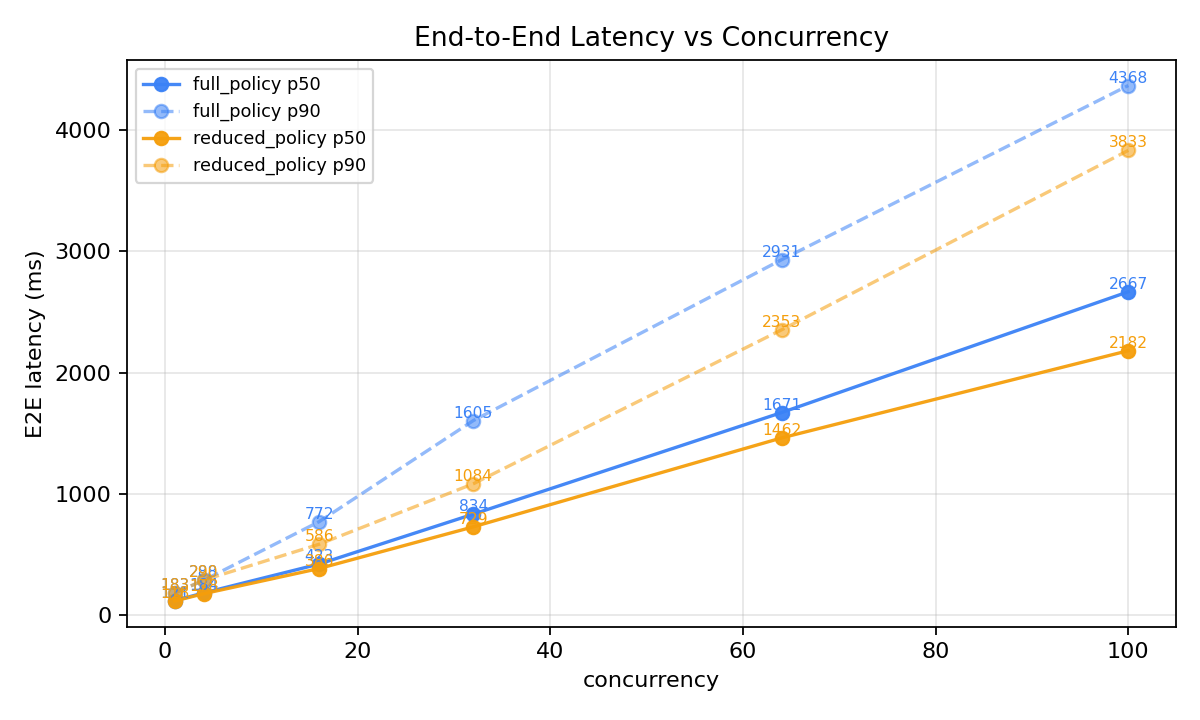}
    \caption{End-to-end latency vs.\ concurrency.}
    \label{fig:a3}
  \end{subfigure}
  \caption{Latency vs.\ concurrency (p50/p90). (a) In the high-concurrency range ($\geq$16)
  the reduced tier's TTFT is stably lower, with the p90 gap exceeding p50 (larger gains on
  tail latency). (b) End-to-end latency improvement grows monotonically with concurrency,
  reaching $\sim\downarrow$18\% p50 at 100.}
  \label{fig:lat}
\end{figure}

\begin{figure}[t]
  \centering
  \begin{subfigure}[b]{0.49\columnwidth}
    \centering
    \includegraphics[width=\linewidth,height=0.78\linewidth,keepaspectratio]{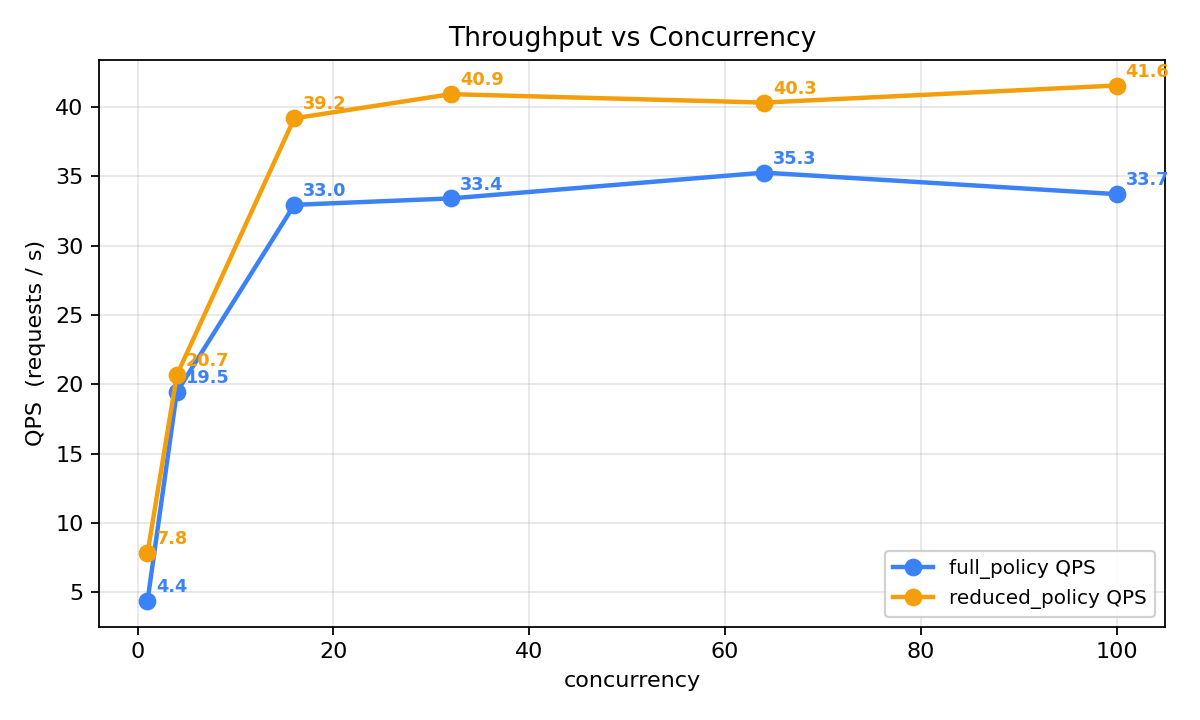}
    \caption{Throughput (QPS).}
    \label{fig:a4}
  \end{subfigure}
  \hfill
  \begin{subfigure}[b]{0.49\columnwidth}
    \centering
    \includegraphics[width=\linewidth,height=0.78\linewidth,keepaspectratio]{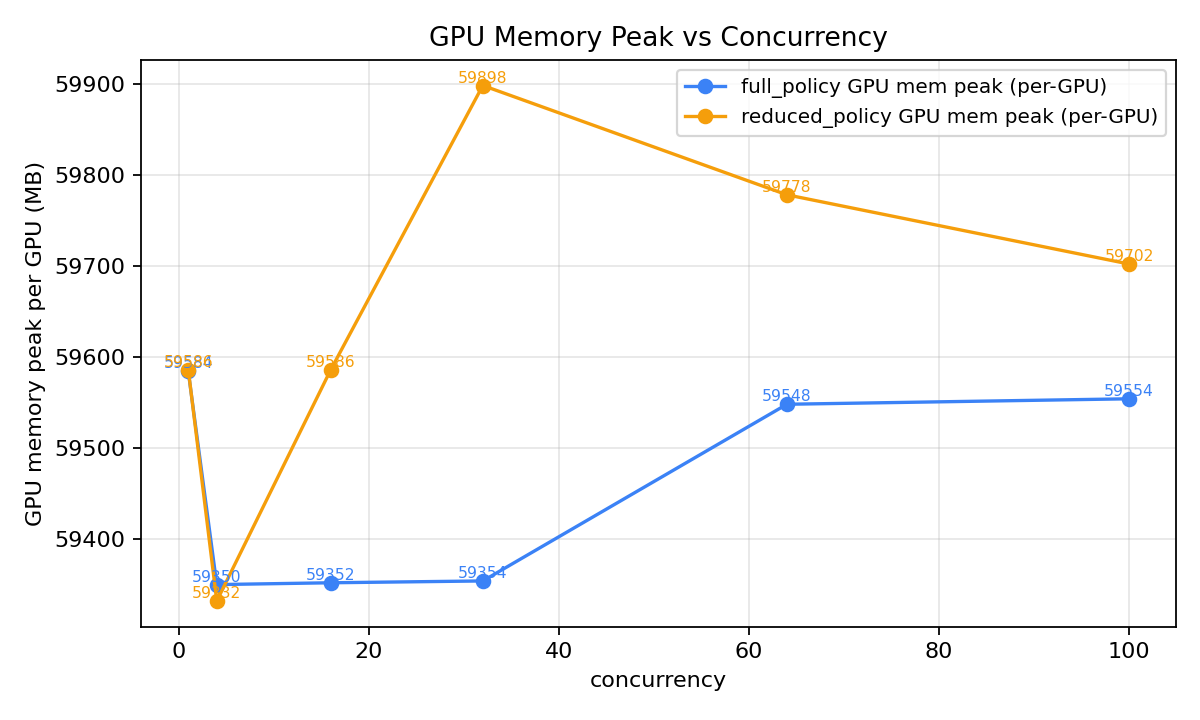}
    \caption{Peak GPU memory (per card).}
    \label{fig:a5}
  \end{subfigure}
  \caption{Throughput and memory vs.\ concurrency. (a) In the high-concurrency range the
  reduced tier's QPS is stably higher ($+14\%$--$+23\%$); the single-concurrency point is
  reference only. (b) Memory is not the bottleneck in this range; the value of reduction shows
  up as a higher concurrency ceiling under the same budget.}
  \label{fig:mem}
\end{figure}

This appendix gives the per-concurrency numbers (Table~\ref{tab:eff_lat}) and the full curves of
the \S\ref{sec:eff} efficiency experiments. The main text reports only a few representative
concurrency points; here we show the full prompt-length distribution (Figure~\ref{fig:a1}) and the absolute curves of
each efficiency metric vs.\ concurrency (Figures~\ref{fig:lat} and~\ref{fig:mem}), to let readers
verify the trend and robustness of the gains.

\begin{table}[h]
  \centering
  \small
  \begin{tabular}{rrrr}
    \toprule
    Concurrency & TTFT & E2E lat.\ (p50) & QPS \\
    \midrule
    1   & $\downarrow$5.6\%  & $\downarrow$2.6\%  & $\uparrow$78.5\% \\
    16  & $\downarrow$26.8\% & $\downarrow$8.8\%  & $\uparrow$18.9\% \\
    32  & $\downarrow$3.6\%  & $\downarrow$12.6\% & $\uparrow$22.5\% \\
    64  & $\downarrow$10.3\% & $\downarrow$12.5\% & $\uparrow$14.3\% \\
    100 & $\downarrow$17.0\% & $\downarrow$18.2\% & $\uparrow$23.2\% \\
    \bottomrule
  \end{tabular}
  \caption{Relative inference-performance improvement of reduced vs.\ full policy at
  different concurrency. TTFT and end-to-end latency are medians (p50). The stable gain of
  prompt reduction appears in the production-relevant high-concurrency range ($\geq$16); the
  single-concurrency QPS is noisy and for reference only.}
  \label{tab:eff_lat}
\end{table}

\paragraph{Setup.} We compare two online tiers: \textbf{full policy} (full policy, mean
single-prompt 8122 tokens) and \textbf{reduced policy} (reduced tier, trimming only redundant
phrasing while keeping all rule details, mean 7093 tokens, $\downarrow$13\%). Both use the
same SIRF-8B model and the same evaluation set ($n{=}1000$). Stress testing runs on SGLang~\citep{zheng2024sglang},
hardware $8\times$L20Y, tensor parallel TP$=$8, \texttt{max\_running\_requests}$=$400,
classification output $\approx$4--5 tokens; concurrency $\in\{1,4,16,32,64,100\}$, 200
requests per tier per concurrency with 5 warmup. Unless noted, TTFT and end-to-end latency are
medians (p50); we also report p90 to reflect the tail. The single-concurrency tier is noisy,
so its QPS is for reference only; production cares about the $\geq$16 high-concurrency range.

Overall, the five figures consistently corroborate the \S\ref{sec:eff} conclusion from four
dimensions (prompt distribution, latency, throughput, memory capacity): a mere 13\% token reduction, with a quantified $4.4$pp B@P95 trade-off while preserving B@P90 and
accuracy (\S\ref{sec:eff}), yields stable high-concurrency latency and throughput gains, and leaves headroom for scaling concurrency.

\section{Interpretability via Logit Lens}
\label{app:lens}

Sections~\ref{sec:main}--\ref{sec:eff} argue SIRF's effectiveness from external metrics
(Black recall, latency and throughput); this appendix gives mechanism-level evidence from
\emph{internal representations}, answering: where exactly does CPT ``learn'' the policy into
the model? Our conclusion: CPT not only fixes the behavior at the final output, but injects
policy-clause-corresponding semantic signals into mid-to-late representation layers; SFT
builds the routing for ``emitting the decision label at the output position.'' Neither alone
suffices; combined, SFT's learned routing lands on CPT's injected, policy-aligned
representations.

\paragraph{Method: Logit Lens.} We probe the model with Logit Lens: project the residual
stream (at the position about to predict the next token, after each transformer block) back
to the vocabulary through the model's own final LayerNorm and lm\_head, obtaining that layer's
vocabulary distribution. Reading the probability of the ``decision-label first-character''
token shows in how many layers and with what strength the correct decision signal emerges. We
use the same dialogue template as in training/inference and set the probe at the first token
of the decision label, ensuring same-distribution with deployment. To verify reliability, we
checked the last-layer prediction against the model's actual inference output, with a 97.6\%
agreement on the decision-label first character, indicating Logit Lens faithfully reflects the
model's real computation path.

\paragraph{$2\times2$ factorial model matrix.} To cleanly separate the contributions of CPT
and SFT, we probe four models the same way: \textbf{Base} (the shared checkpoint \texttt{Qwen3-VL-8B-Instruct}, \S\ref{sec:train}),
\textbf{CPT} (Base$+$CPT), \textbf{SFT} (Base$+$SFT), \textbf{SIRF} (Base$+$CPT$+$SFT, main model). The
target tokens for decision labels and the policy-keyword list are pre-registered in a config
before the experiment, forbidden to be adjusted post-hoc, to rule out cherry-picking.

\begin{figure}[t]
  \centering
  \includegraphics[width=\columnwidth]{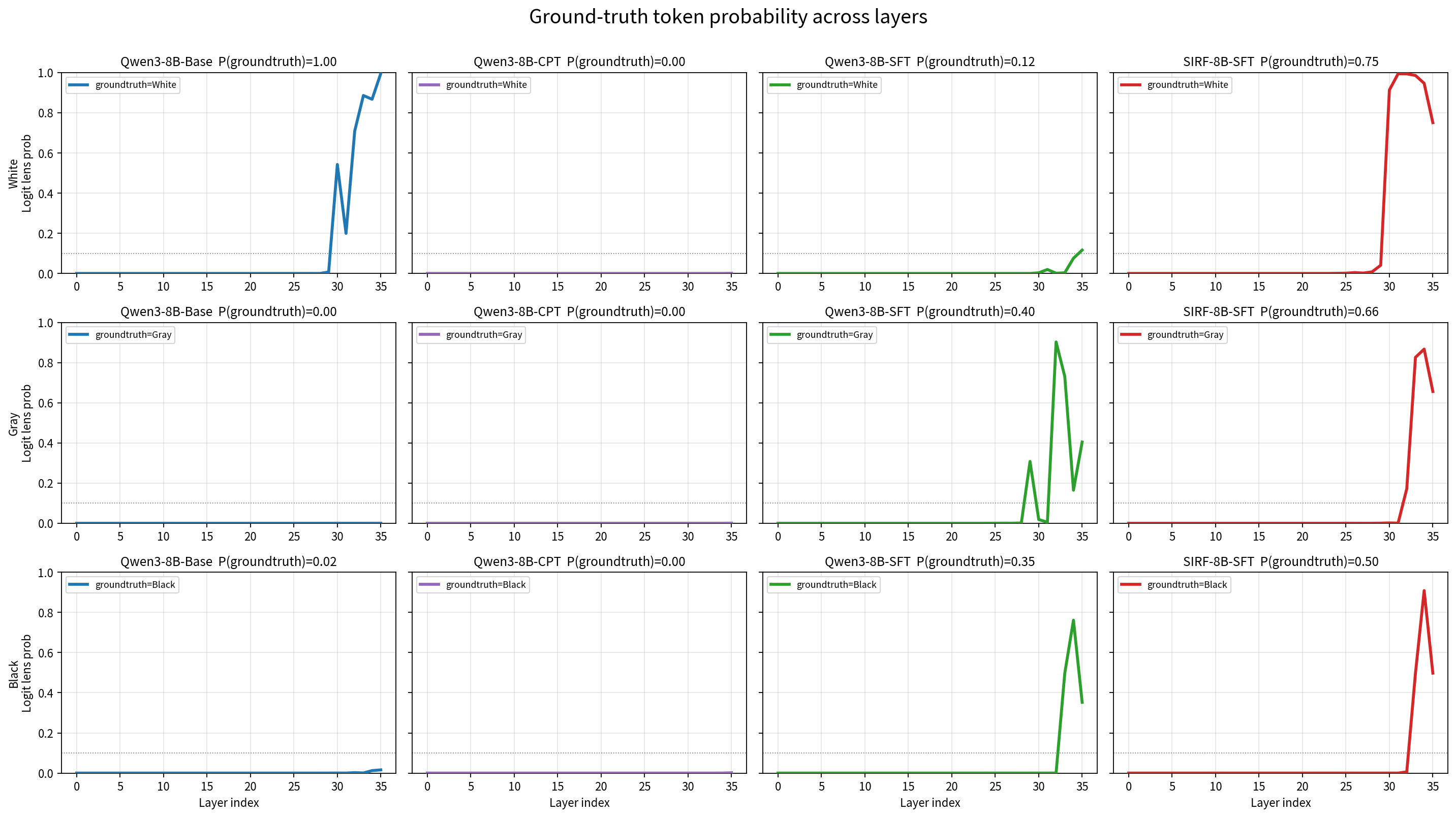}
  \caption{Ground-truth label probability across layers for three representative samples and
  four models. Rows: true label (White, Gray, Black); columns: model (Base, CPT, SFT, SIRF).
  Each subplot plots the ``correct-label first character'' probability under Logit Lens
  per layer; the dashed line is the emergence threshold 0.1, and the title's P(groundtruth) is
  the final-layer probability.}
  \label{fig:lens}
\end{figure}

Figure~\ref{fig:lens} reveals three key phenomena:
\begin{enumerate}
\item \emph{CPT improves signal ``strength,'' not ``earlier appearance.''} In the Gray and
  Black rows, the correct-label signal for both SFT and SIRF emerges near L30, but SIRF's
  final-layer probability is markedly higher (Gray: 0.66 vs 0.40; Black: 0.50 vs 0.35). So
  CPT's contribution is not earlier emergence but stronger, more robust signal after
  emergence, consistent with the intuition that CPT injects more stable semantic
  representations.
\item \emph{CPT alone does not emit labels (``knowledge but no routing'').} The CPT column
  has near-zero correct-label probability throughout, with its final prediction often on
  irrelevant tokens such as markdown title symbols. This shows CPT changed the model's
  knowledge and expression preferences but did not build the structural mapping of
  ``outputting the decision-label first character at the assistant start''; this routing is
  learned only by SFT.
\item \emph{SFT builds routing but is easily hijacked by surface cues; CPT$+$SFT is stable.}
  The SFT column gives the correct label at the end but with lower probability and is
  sometimes biased to a wrong label by surface cues; the SIRF column (CPT$+$SFT) gives the
  highest, cleanest correct signal on all three classes. This is the internal mechanism
  corresponding to SIRF recalling more risk than SFT at the high-precision operating point in
  \S\ref{sec:main}.
\end{enumerate}

\paragraph{Caveats.} Logit Lens reflects only the token probability after projecting the
residual to the vocabulary; it is evidence of CPT-induced layer-wise emergence rather than a
strict ``feature localization,'' and the decision signal is approximated by the label first
character (a few labels sharing a first character have slight confusion). These do not affect
the above cross-model relative comparison, but validation with larger scale and stronger
causal methods is future work.

\section{Per-Component CPT Ablation}
\label{app:ablation}

Table~\ref{tab:ablation} is the full version of the ablation summarized in \S\ref{sec:ablation},
together with the training recipe shared by all arms.

\begin{table}[h]
  \centering
  \footnotesize
  \setlength{\tabcolsep}{3pt}
  \begin{tabular}{lrrr}
    \toprule
    CPT corpus & B@P95 & B@P90 & M-F1 \\
    \midrule
    no CPT (Qwen3-8B-SFT)      & 56.2 & 83.4 & 81.9 \\
    domain-only (no policy)    & 55.7 & 83.6 & 80.4 \\
    EntiGraph$+$MAGA only      & 64.6 & 83.2 & 80.7 \\
    CoT only                   & 69.5 & 83.6 & 80.9 \\
    w/o CoT                    & 70.6 & 81.2 & 79.7 \\
    \textbf{full CPT (SIRF-8B)}& \textbf{71.3} & \textbf{84.7} & 80.4 \\
    \bottomrule
  \end{tabular}
  \caption{Per-component CPT ablation ($n{=}1000$): the CPT corpus is varied while the 8B base, the
  SFT data and schedule, and the evaluation set are held fixed. End rows are the
  Table~\ref{tab:main} runs. Since Recall@P95 is a step quantity (\S\ref{sec:setup}), adjacent
  middle rows are not a precise ranking.}
  \label{tab:ablation}
\end{table}

\paragraph{Training details.} All arms in Table~\ref{tab:ablation} use the identical
recipe: CPT with next-token prediction, full-parameter, 1 epoch, LR $1\text{e-}5$, cosine
schedule, warmup $0.03$, cutoff 4096 with packing; then SFT, full-parameter, 3 epochs, LR
$1\text{e-}5$, cutoff 16384, on the same $40{,}989$-sample set, with an effective batch size of
512 in both stages. The CPT blocks are: account-level CoT $22{,}674$ documents ($\sim$51M
characters), EntiGraph$+$MAGA $10{,}041$ ($\sim$17.5M), public fraud corpus $59{,}104$
($\sim$7.1M), general anti-forgetting corpus $9{,}850$ ($\sim$10.5M).

\section{Prompt and Feature Serialization}
\label{app:prompt}

The account data are proprietary and cannot be released, and the exact feature schema is
business-confidential, so we specify the input \emph{structure} rather than the individual fields.
The prompt is \texttt{\{policy\}$\backslash$n\{features\}} and the target is the verdict label
alone. The feature block serializes a heterogeneous account view into a fixed sequence of
bracketed sections, one per signal domain --- account profile, published content and media,
interaction traces, and abuse-report signals --- and within each section one line per field, in a fixed order, of the form
\texttt{- <field>: <value>}. Recent content is rendered as a
title/body pair per item, with explicit markers for missing or restricted items.

Three conventions matter more for reproduction than the field list itself. Empty domains are
rendered explicitly as \texttt{empty} rather than dropped, so the section layout is identical
across accounts and the model can distinguish ``no signal'' from ``field absent''. The recent-content
window is fixed-length (the most recent 10 items) rather than token-budgeted, which is what keeps
prompt length roughly stable (mean $\sim$12.1k characters with the full policy, of which
$\sim$10.7k is the policy). All identifiers are removed or pseudonymized before serialization, and
internal-ecosystem labels are stripped so that no system label can leak through the feature block
(\S\ref{sec:corpus}).